# Subject islands do not reduce to construction-specific discourse function

*Mandy Cartner[a], *Matthew Kogan[b], *Nikolas Webster[b], Matthew Wagers[b], Ivy Sichel[b]

[a]Tel Aviv University, 55 Chaim Levanon St, Tel Aviv-Yafo, 6997801, Israel

[b]University of California, Santa Cruz, 1156 High St, Santa Cruz, CA 95064, United States

*co-first authors

A central question about our shared capacity for language is how it is integrated with other cognitive systems. One important debate focuses on the extent to which the form of linguistic expressions is grounded in their communicative function: Can all constraints on linguistic form be attributed to the way constructions package information, or is linguistic form autonomous of meaning and function? One area of disagreement involves *islands*: phrases which block the formation of long-distance filler-gap dependencies (Ross, 1967). Grammatical subjects are considered islands, since questioning a sub-part of a subject results in an ill-formed sentence, e.g., “Which topic did the article about inspire you?”. Autonomous syntactic approaches to islands attribute this ungrammaticality to the abstract *movement* dependency between the *wh*-phrase and the subject-internal position with which it is associated. An alternative developed in Abeillé et al. (2020) suggests that subjects’ island status is specific to the information structure of *wh*-questions, suggesting that subjects are not islands for *movement*, but for *focusing*, due to their discourse-backgroundedness. This predicts that other constructions that involve movement but not focusing should not create a subject island effect. We test this in three acceptability studies, using a factorial design to isolate subject island violations across three constructions: *wh*-questions, relative clauses and topicalization. We find a subject island effect in each case,

Corresponding author: mandycartner@mail.tau.ac.il

despite only *wh*-questions introducing what Abeillé et al. (2020) call "a clash in information structure". We argue that this motivates an account of islands in terms of syntactic representations shared across constructions, independent of communicative function.

## 1. Introduction

Every language user possesses a largely unconscious capacity to form well-structured sentences from an infinite range of possibilities. We know not only which word orders are acceptable in our language, but also which rearrangements sound wrong— even when the intended meaning is clear. The nature of the system of rules and representations that comprises this knowledge is a source of intense theoretical debate, particularly concerning its level of abstractness: Is all knowledge concerning the form of linguistic expressions attributable to functional properties, such as an expression's intended meaning or use, or is there reason to posit a dedicated system of rules, independent of communicative function, responsible for the structure of language?

A long standing tradition in the study of language is the usage-based approach, which highlights language's communicative function, and views grammar as a set of *constructions*, defined by different combinations of form, meaning, and function, such as *wh*-questions, relative clauses, topicalization structures, cleft constructions, and passives (e.g., Abeillé et al., 2020; Ambridge & Goldberg, 2008; Engdahl, 1997; Erteschik-Shir, 1973; Goldberg, 1995; Goldberg, 2006; Kuno, 1987; Tomasello, 1998). Within this view, grammatical knowledge is a network of form-meaning pairs, induced from the input via general cognitive processes. Abstract grammatical notions, such as 'subject', are not theoretical primitives, and are understood, instead, to emerge from generalizations over recurring patterns in language use.

Usage-based approaches are in apparent tension with the discovery of locality restrictions called *islands* (Ross, 1967) and their characterization in the generative tradition as abstract, structural rules. The term 'islands' refers to a set of environments which block the possibility of a filler-gap dependency, a general term for configurations in which some phrase appears in a clause-initial position but is associated with another, more embedded position, in which it is interpreted along with the predicate. We will refer to the phrase in initial position as the *filler* ('who' in (1)), to the underlined position it relates to as a *gap*, and to the relation between the two as a *filler-gap dependency*. The dependencies in (1) are incrementally longer due to iterative subordination of the embedded clause which hosts the gap in (1b-c), and they are all grammatical. This type of dependency has been considered unbounded, because it can span an indefinite number of embedded complement clauses.

(1) a. Who did Jaden see ___?

b. Who did Aidan think that Jaden saw ___?

c. Who did Mariella say that Aidan thinks that Jaden saw __ ?

At the same time, there are embedding domains which exclude a filler-gap dependency. Two examples of islands include adjunct clauses, illustrated in (2), and subject phrases, in (3). In each of these examples, the embedding domain is marked with brackets. The (a) examples give a sentence without a filler-gap dependency, and the (b) examples give the corresponding sentence with a filler-gap dependency that terminates in the target domain. We will refer to a filler-gap dependency into a designated phrasal domain as 'sub-extraction' from that domain: the example in (2b) features sub-extraction from an adjunct, in (3b) sub-extraction from a subject, and in (4b) from an object. Unlike the embedded clausal complements shown in (1), sub-extraction from adjuncts and from subjects is distinctly unacceptable. This pattern of acceptability provides evidence that adjunct clauses and subjects are *islands* for filler-gap dependencies. However, not every complex domain is an island: the same phrase that is an island in subject position (3b), allows a filler-gap dependency in object position (4b).

(2) a. Jaden meditated [ before meeting Mariella ].

b.*Who did Jaden meditate [ before meeting ___ ] ?

(3) a. [ A friend of Jaden ] invited Mariella to the party.

b.*Who did [ a friend of ___ ] invite Mariella to the party?

(4) a. Mariella invited [ a friend of Jaden ] to the party.

b. Who did Mariella invite [ a friend of __ ] to the party?

These examples show that one kind of filler-gap dependency, namely *wh*-question formation (WHQ, henceforth), cannot span an island, i.e., cannot involve a dependency that reaches into a subject or an adjunct. The significant discovery of Ross (1967) was that many other types of filler-gap dependencies beside WHQs are sensitive to islands. For example, relativization (RC, henceforth) (5) and topicalization (TOP, henceforth) (6) show similar subextraction restrictions.

(5) a. This is the guy [ who [ Mariella invited [a friend of ___ ] to the party] ].

b.*This is the guy [ who [ [a friend of ___ ] invited Mariella to the party] ].

(6) a. That guy, Mariella invited [ a friend of ___ ] to the party.

b. *That guy, [ a friend of ___ ] invited Mariella to the party.

These constructions undeniably vary along several syntactic, semantic, and pragmatic dimensions. But their shared sensitivity to the presence of islands has suggested that the variation among these constructions is less important— for island locality— than what they share. The element that is sensitive to island locality is the filler-gap dependency common to all of these constructions. This discovery set in motion a series of departures from the traditional view of grammar as a set of constructions (shared by early generative frameworks, such as Chomsky 1965), and towards a view of syntax as autonomous of meaning and function, in which what is sensitive to the constraints of locality is an abstract syntactic representation of a filler-gap dependency. In particular, the Subject Condition (Cattell, 1976; Kayne, 1981; Chomsky, 1973; Huang, 1982; Pesetsky, 1982; Privoznov, 2021; Ross, 1967) asserts that constituents within a syntactic subject cannot form a dependency with a filler in the main clause. While the particular formulations of the Subject Condition differ, they all share the claim that all of these constructions are sensitive to the Subject Condition.

Alongside adjunct islands (2) and subject islands (3), a number of other island domains have been identified, such as embedded *wh*-questions (called *wh*-islands). Subject islands and adjunct islands are grouped together as strong islands, which are domains that categorically disallow any type of filler-gap dependency. *Wh*-islands, on the other hand, are examples of weak islands, which are domains that block filler-gap dependencies selectively based on the syntactic and semantic status of the filler. Strong and weak islands have come to be analyzed differently: the selective opacity of weak islands is argued to be best accounted for by a semantic or pragmatic analysis (e.g., Abrusán, 2014; Cinque, 1990; Kiss, 1993; Szabolcsi & Zwarts, 1990, 1993; Szabolcsi & Lohndal, 2017), while the absolute opacity of strong islands is accounted for configurationally, as a function of their position in the hierarchical structure (Cattell 1976; Huang 1982; Kayne, 1981, 1994; Uriagereka, 1999; Johnson, 2003; Privoznov, 2021). Our empirical focus in this paper is subject islands, a variety of strong islands, defined structurally and syntactically within generative approaches.

A line of research, pioneered in Erteschik-Shir (1973), challenges the claim that the source of (un)acceptability for certain types of filler-gap dependencies is purely syntactic[1]. Instead, this approach assigns a significant role to discourse-based constraints rooted in information structure (henceforth, IS) concepts such as backgroundedness, focus, and prominence (e.g., Engdahl, 1997; Goldberg & Ambridge, 2008; Kuno, 1987), as these categories apply to types of dependencies or to the extraction domain (Ambridge & Goldberg, 2008; Engdahl, 1997; Erteschik-Shir & Lappin, 1979; Kuno, 1987; Lu et al., 2024; Namboodiripad et al., 2022). In this paper, we respond to one particular version of this approach, which we label the *constructional IS profile* theory, and which, as its name suggests, distinguishes between RCs, WHQs, and other dependencies based on their IS profiles (Abeillé et. al., 2020; Winckel et. al., 2025). We address the specific claim that IS-based notions such as backgroundedness interact with the IS profile of the filler, a claim that was recently argued for in Abeillé et. al. (2020) and Winckel et. al. (2025). Based on experimental findings that sub-extraction of a prepositional phrase (PP) is rated less acceptable out of subjects vs. objects in WHQ constructions, but not in RCs, Abeillé et al (2020) propose that unacceptable sub-extraction out of a subject arises from a "clash" formalized as the Focus Background Constraint (FBC): "a focused element should not be part of a backgrounded constituent." This analysis draws a clear line between WHQs and RCs, and hinges critically on an assumption that subjects are typically backgrounded (i.e. discourse familiar, presupposed, unfocused; see Section 2 below), and objects/post-verbal constituents are part of the focus. WHQs place an element in a focused position, a position associated with the introduction of discourse-new material, whereas RCs specify or attribute a property of the RC head. From this perspective, sub-extraction from a subject is predicted to be unacceptable for WHQs, but not RCs, because the filler is focused in a WHQ construction, and this conflicts with its direct relation to a backgrounded extraction domain (i.e. a subject). Abeillé et. al. (2020) motivate the FBC on the basis of five acceptability judgment studies in English and French, both of which bear out the predicted contrast between WHQs and RCs.

[1] Syntactic approaches to strong islands do not deny that discourse function plays a role, but it is mediated through syntactic structure, and the account is ultimately syntactic (e.g., Fiengo & Higginbotham, 1981; C.-T. J. Huang, 1982; Kush et al., 2018; N. Huang et al., 2025). What differs in the trajectory of research stemming from Erteschik-Shir (1973)'s work is the development of an alternative hypothesis that fully reduces island effects to non-syntactic factors, in which case large and useful simplifications could be made to the model of the grammar.

Crucially, the observed difference between WHQs and RCs in English holds only when an entire prepositional phrase (PP) is subextracted— a configuration called "pied-piping"— but not when just the noun phrase is subextracted, leaving a preposition at the gap site— called preposition stranding or "p-stranding"— as in (3-4). However, there is no known IS contrast between the two sub-extraction strategies. For this reason, it remains unclear why the FBC, articulated in terms of the discourse functions of the filler position and gap site, should be sensitive to the syntactic category of the fronted constituent. Although languages differ in the availability of sub-extraction with pied-piping and p-stranding (e.g. Romance, which only permits pied-piping, and English, which permits both pied-piping and p-stranding), we argue that p-stranding provides a more natural testing ground for investigating subject islands across constructions in English. Despite prescriptive pressure against its use, p-stranding is widely acceptable in both formal and informal English registers across a number of syntactic environments (Huddleston & Pullum, 2002). P-stranding also provides an unambiguous cue to the position of the gap, which is a challenge in resolving any filler-gap dependency. Moreover, previous research has argued that pied-piping and p-stranding are distinct filler-gap dependencies, with only the latter exhibiting movement that feeds islandhood (Bianchi & Chesi 2014, 2015; Jurka, 2010). For these reasons, we investigate sub-extraction in English using p-stranding.

We further investigate the predictions of the FBC by probing the acceptability of sub-extraction from subjects across WHQ and RC, as in Abeillé et al (2020), and we add topicalization (TOP) to the comparison. We argue against the FBC by showing that the predictions generated by the FBC fail for TOP dependencies in addition to RCs. For TOP, the FBC predicts no subject island effect, since here part of the backgrounded subject constituent is associated with a topicalized, backgrounded constituent, and should not produce the IS clash that underpins subject island effects. To test these predictions, we conducted three separate large-scale acceptability studies (WHQ, RC, and TOP). Each utilized a factorial design manipulating Position (extraction domain, *subject* or *object*), DP Complexity (*simple* or *complex*), Extraction Type (*no*, *full*, or *sub-extraction*) to estimate the super-additive cost of sub-extraction (Sprouse, 2007; Sprouse et al., 2012). This factorial design allows us to clearly define an island effect, by factoring out independent variables which may influence the acceptability of sentences containing island violations, such as DP complexity and extraction. We operationalize the subject island effect as the difference between the costs of subject and object sub-extraction, while factoring out the potential contribution of DP complexity and full DP extraction from the acceptability of sub-extraction in each position.

Across all three experiments, we found evidence for a subject island effect: with TOP, WHQ, and RC dependencies, the observed cost of sub-extraction was greater than the combined cost of DP complexity and extraction, and this sub-extraction cost for subject positions consistently exceeded that found for object positions. This indicates the presence of an additional penalty associated with sub-extraction from subjects that is not predicted by DP complexity or extraction, nor is it shared by object sub-extraction. These results corroborate existing, cross-linguistic findings which have similarly provided evidence for subject island effects across

WHQ, RC, and TOP dependencies (Kobzeva et al., 2022 for WHQ and RC in Norwegian; Kush et al., 2018, 2019 for both WHQ and RC, and WHQ and TOP, in Norwegian; Sprouse et al., 2016 for WHQ and RC in English, although variance is reported for Italian). Building upon this background, the cross-constructional analysis presented in Section 7 provides evidence of a stable and invariant subject island effect (i.e. the difference between sub-extraction from subject and object positions), despite substantial variation in the absolute acceptability of extraction across the three constructions. The present study examines all three constructions within the same language. Together with these previous cross-linguistic findings, our results militate against a purely discourse-based account of subject islandhood.

Next, in Section 2, we introduce discourse-function based theories of island violations and the predictions of the FBC in more detail before presenting our super-additive experiment design in Section 3, which we use to test subject island effects in WHQ (Experiment 1), RC (Experiment 2), and TOP (Experiment 3), respectively.

## 2. Discourse function-based accounts

Beginning with Erteschik-Shir (1973), the guiding intuition of discourse function-based accounts of islandhood has been that the backgrounded (or presupposed) nature of the extraction domain is responsible for the degraded status of filler-gap dependencies. Roughly speaking, a constituent is backgrounded if it is not part of what the sentence asserts or presents (Cuneo & Goldberg, 2023). Under these accounts, the source of islandhood is located not in the syntax, but in IS, which tracks distinctions between what is backgrounded in a given sentence vs. what is asserted or in focus.

Erteschik-Shir (1973) argues for a constraint on sub-extraction tied to a notion of "semantic dominance," which can be understood as the "at issue" content of a given utterance. For example, Erteschik-Shir (1973) notes that even clausal complements, which are generally not islands for dependency formation, can sometimes block a filler-gap dependency depending on whether the matrix verb conveys that its clausal complement is presupposed (cf. 7 vs. 8). A verb like *rejoice* typically signals that the content of its embedded clause is presupposed to be true and thus backgrounded in context, as it is in (8a). In contrast, the content of the embedded clause in (7a) is not necessarily presupposed, and sub-extraction is possible.

(7) a. Nora said that Marcus visited his mother.

b. Who did Nora say that Marcus visited ___ ?

(8) a. Nora rejoiced that Marcus visited his mother.

b. *Who did Nora rejoice that Marcus visited ___ ?

Erteschik-Shir (1973) therefore argues that filler-gap dependency formation is blocked across an extraction domain that carries backgrounded information. Discourse function accounts in the spirit of Erteschik-Shir (1973) argue in favor of the strongest version of this hypothesis, where *all* filler-gap dependencies— WHQs, RCs, TOPs, etc.— interact similarly with backgroundedness (Cuneo & Goldberg, 2023; Erteschik-Shir, 1982; Goldberg, 2006, 2013; Goldberg et al., 2024; Lu et al., 2024; Namboodiripad et al. 2022). This is expressed in the formulation of the BCI constraint from Cuneo & Goldberg (2023) in (9) below. We refer to this family of theories as *direct backgroundedness* approaches.

(9) Backgrounded Constructions are Islands (BCI):

> Constructions are islands to [filler-gap] dependency constructions to the extent that their content is backgrounded within the [extraction] domain of the [filler-gap] dependency construction.

Cuneo & Goldberg (2023) further clarify their formulation of (9) with the following: they expect acceptability to be gradient, and 'island status' is anticipated to vary with the degree of backgroundedness associated with the extraction domain. To the extent that all sentences include constituents that are more backgrounded and less backgrounded, it follows that any constituent that contains a gap has the potential to elicit an island violation under the *direct backgroundedness* view, depending on its degree of backgroundedness. Cuneo & Goldberg (2023)'s conception of backgroundedness is as a gradient notion which depends on particular lexical choices, scope configuration, and other factors. In an investigation into a wide array of construction types with filler-gap dependencies and whether the acceptability of these constructions correlates with measures of backgroundedness, Cuneo & Goldberg (2023) found that their filler-gap dependency stimuli tended to be more acceptable if baseline items corresponding to the extraction domain were independently evaluated as "less backgrounded" in two other experimental tasks (though see Momma & Dillon, 2023 for arguments against the conclusion that island status is *causally* related to assessments of backgroundedness).

*Constructional IS profile* theories predict that the acceptability of a filler-gap dependency is dependent on the interaction between the IS status of the extraction domain and the IS status of the filler, not simply based on the IS of the extraction domain itself, as in *direct backgroundedness* approaches. *Constructional IS profile* approaches distinguish between RCs, WHQs, and other filler-gap dependencies, based on each construction's IS profile (Abeillé et al., 2020; Winckel et al., 2025). In these accounts, island status is claimed to vary according to the discourse functions that particular constructions impose on their sub-parts. More concretely, unacceptability arises due to a clash in discourse functions between the extraction domain and the functions of a particular construction. It is relevant to note that the BCI, as articulated in Cuneo & Goldberg (2023), operates according to a similar logic: unacceptability is due to a “clash of functions” in foregrounding a constituent with a long-distance dependency construction and simultaneously backgrounding the constituent according to the semantic-pragmatic properties of the ‘base’ domain (i.e. extraction domain). However, the characterization of the dependency varies across the two frameworks, leading to distinct predictions. For *constructional IS profile* theories, which we outline in more detail below, clashes may or may not arise due to the discourse functions of the long-distance dependency construction and the extraction domain (and more precisely, the discourse function that the construction imposes on the extracted constituent); for the BCI, all long-distance dependency constructions are assumed to foreground (or make prominent) a constituent.

Abeillé et al. (2020) formulate the FBC (10) to account for their findings of improved acceptability for PP sub-extraction out of subjects in RCs in comparison to WHQs. The application of the FBC is further restricted in Winckel et al. (2025) to only apply to filler-gap constructions, as in (11), but the essence of the FBC is maintained across both papers: a filler (an extracted element) that is more focused than the constituent that contains its gap will result in an unacceptable construction.

(10) FBC (Abeillé et al., 2020)

A focused element should not be part of a backgrounded constituent.

(11) FBC revised (Winckel et al., 2025)

> An extracted element should not be more focused than its (non-local) governor. Hence the greater the difference in focus between a focused element and its less focused governor, the more infelicitous the dependency will be.

The logic of the FBC is relational, tying (un)acceptability to the (in)compatibility of discourse functions of both the filler-gap dependency construction and the extraction domain. Therefore, the predictions of the FBC vary across individual constructions. In WHQs, the extracted element (i.e. the *wh*-phrase) is the focus, characterized as containing prominent or "at-issue" content which is otherwise unpredictable or nonrecoverable from the utterance, standing in contrast to the backgrounded content of an utterance (Gundel & Fretheim, 2006; Lambrecht, 1994)[2]. Thus, the FBC predicts that sub-extraction of a *wh*-phrase from a backgrounded constituent, such as a subject, will result in unacceptability, but for RCs the FBC makes a different prediction. Since RCs apply some property to an entity (Kuno, 1976) without necessarily specifying a discourse function, they are compatible with backgroundedness, topicality, or focus (Gundel, 1988; Lambrecht, 1994). Therefore a gap contained in a grammatical subject that is linked to a filler fronted via relativization will not engender a clash according to the FBC. Thus, the FBC predicts a contrast between WHQs and RCs: Only the former should induce a subject island effect. This is indeed what Abeillé et al (2020) report, at least for constructions that involve pied-piping (i.e., when a preposition is fronted alongside the filler).

The FBC makes similar predictions for TOP, a construction not tested in Abeillé et al. (2020). TOP involves the extraction of a topic constituent, characterized as an established matter of concern, about which new information is added (Lambrecht, 1994; Reinhart, 1981; Strawson, 1964). This construction serves to mark the extracted element as already backgrounded in the discourse and to mark the corresponding proposition as being about this referent (Lambrecht, 1994; Prince, 1983, 1984). The final sentence in (12) uses TOP felicitously to refer to the backgrounded constituent *the witch*, but the final sentence of (13), which uses an indefinite determiner to introduce a novel discourse referent, is infelicitous in the TOP construction.

[2] Focus is often defined in terms of discourse newness/givenness (Schwarzschild, 1999), whereby focus is taken to mark new information in opposition to the presupposed or given information in an utterance. However, equating focus with newness can be misleading (Lambrecht, 1994), a conclusion which finds amplified support in recent reading time studies (Hoeks et al., 2023).

(12) Once there was a wizard and a witch. The wizard had two sons. The first was tall and brooding. The second was short and vivacious. *The witch, the wizard had been madly in love with ___.*

(13) Once there was a wizard. The wizard had two sons. The first was tall and brooding. The second was short and vivacious. **A witch, the wizard had been madly in love with ___.*

Given the discourse function of TOPs, the FBC predicts no clash in discourse function for TOP sub-extraction from a subject, unlike WHQs, leading to the expectation that subjects should not be islands for TOPs. Our study is designed to test the FBC predictions systematically across a range of constructions in English. We are inspired by counterevidence to the FBC in prior experimental work which observed robust island effects for TOP and WHQs and separately for RCs and WHQs in Norwegian (Kobzeva et al., 2022; Kush et al., 2018, 2019). They provide an account that acknowledges the empirical differences between these types of dependencies within a syntactic perspective on islands that does not implicate constructions as the basic unit of analysis. We note that Sprouse et al. (2016) as well report evidence of island effects in WHQs and RCs in English. We situate our work amidst these findings, drawing inspiration from the methodological design developed by Sprouse (2007) and Sprouse et al. (2012) to test islands experimentally, uniting all three dependency types (RC, WH, and TOP) under one language, and narrowing the focus to subject islands, for a direct response to Abeillé et al. (2020).

In three large-scale acceptability judgment experiments we address the specific claim from Abeillé et al. (2020) and Winckel et al. (2025) that IS-based notions such as backgroundedness directly predict the acceptability of extraction, via an interaction of the IS profile of the construction with the IS profile of the filler. We investigate the FBC by probing the acceptability of subject sub-extraction across three constructions: WHQs and RCs, as in Abeillé et al. (2020), to which we add TOPs. The FBC predicts subject island violations to only arise for WHQs, but not for TOP constructions nor for RCs. We argue that the source of subjects' islandhood cannot be an IS clash by showing consistent island effects in WHQs and RCs with p-stranding, as well as in TOP, suggesting that subjects' islandhood reflects a constraint on the abstract mechanism that forms a filler-gap dependency, shared by WHQs, RCs, and TOPs alike.

## 3. Our study

The present study tests the cost of sub-extraction out of subjects across three constructions (WHQs, RCs, TOP), each with different IS profiles. All three of the experiments presented in this paper utilize a factorial design that crosses the two positions of the potential filler-gap site (*Subject* and *Object*) with five conditions, defined below (see Table 1), that allow us to isolate the specific potential "island cost". We created these conditions following the strategy first introduced in Sprouse (2007) and Sprouse et al. (2012), designed to isolate the effect of an island violation on the acceptability of a sentence by factoring out other properties of island constructions that independently affect acceptability: DP complexity and filler-gap dependency length. This methodology has been proven to be an effective tool to probe for the presence of islands across many languages and constructions (e.g., Almeida, 2014; Keshev & Meltzer-Asscher, 2017; Kobzeva et al. 2022, Kush et al., 2018, 2019; Pañeda et al. (2020); Sprouse et al., 2016; Stepanov et al., 2018; Tucker et al., 2019; Vincent et al., 2022).

DP complexity refers to the relative complexity of the structure of a DP: for the purposes of this study, what matters is whether a DP also contains a PP complement— for example, "the driver" is a simple DP, whereas "the driver of the car" is a complex DP. Complex DPs have been found to be dispreferred both generally as well as in subject position and/or sentence initial positions (e.g., Rizzi & Shlonsky, 2006), leading to lower ratings of acceptability between sentences with simple DP subjects and sentences with complex DP subjects. Since subject islands by definition involve a gap within a grammatical subject, all sentences with sub-extraction from subjects will have complex DP subjects, contributing to the sentence's overall acceptability.

Filler-gap dependency length is also well-known to affect both ease of processing and overall sentence acceptability, such that longer filler-gap dependencies are relatively degraded (Holmes & O'Regan, 1981; King & Just, 1991; Gibson, 1998). This is a crucial factor to consider when measuring subject sub-extraction compared to object sub-extraction: Subject sub-extraction involves a shorter linear dependency in English, which could independently improve its acceptability compared to object sub-extraction, and this may obscure the subject island effect. Studies show that dependencies formed between a filler and a subject gap, such as the Subject RC in (14), are easier to process and more acceptable than dependencies formed between a filler and an object gap, as in the Object RC in (15) (e.g., Holmes & O'Regan, 1981; Traxler et al., 2002).

(14) I noticed the investigator [ that _ had already questioned the driver ].

(15) I noticed the driver [ that the investigator had already questioned _ ].

This effect, known as the *subject advantage*, has been attributed to a variety of factors, including increased pressure on working memory resources in object RC configurations, differences in dependency length and depth of embedding of the extracted constituent, and relative distributional frequencies of subject and object RCs, among others; see Lau & Tanaka (2021) for a recent overview. Due to the inherent asymmetry between subject and object extractions, this independent contrast must be factored out in order to measure the subject island effect correctly.

To understand the effect of a subject island violation on the acceptability of a sentence, then, it is not enough to simply compare a sentence with a gap inside of a subject, such as (16) below— to one with a gap within an object, like (17), as Abeillé et al. (2020) do. For example, there may be an independent cost of complex DPs in subject position which, when added together with the dependency cost, could lead to predictably lower ratings based exclusively on these two costs and unrelated to the existence of a specific subject island.

(16) Which crime did Stephanie explain [$_{\text{SUBJ.}}$ the investigator of _ ] had already questioned the driver?

(17) Which car did Stephanie explain the investigator had already questioned [$_{\text{OBJ.}}$ the driver of _ ] ?

Our design, illustrated with a sample itemset in Table 1, allows us to isolate an acceptability cost for DP complexity in each position by comparing across *Complex* vs. *Simple* conditions (Subjects: A - C; Objects: A - B), and to isolate a cost for dependency length in each position by comparing across *No Extraction* vs. *Full Extraction* conditions (Subjects: A - E; Objects: A - D). The simple DP condition serves as a baseline for both subjects and objects, as both of these arguments are simple DPs. Additionally, we note that sub-extraction is only possible for complex DPs, and thus we cannot fully cross Extraction Type and Complexity factors.

| | | |
|---|---|---|
| | *No Extraction* (baseline declaratives) | |
| A | Simple | Stephanie explained the investigator had already questioned the driver. |
| B | Complex object | Stephanie explained the investigator had already questioned the driver of the car. |
| C | Complex subject | Stephanie explained the investigator of the crime had already questioned the driver. |
| | *Simple DP, Full Extraction* | |
| D | Object | Which driver did Stephanie explain the investigator had already questioned _ ? |
| E | Subject | Which investigator did Stephanie explain _ had already questioned the driver? |
| | *Complex DP, Full Extraction* | |
| F | Object | Which driver of the car did Stephanie explain the investigator had already questioned _? |
| G | Subject | Which investigator of the crime did Stephanie explain _ had already questioned the driver? |
| | *Complex DP, Sub-Extraction* | |
| H | Object | Which car did Stephanie explain the investigator had already questioned the driver of _? |
| I | Subject | Which crime did Stephanie explain the investigator of had already questioned the driver? |

**Table 1**. Example set from Experiment 1.

A sub-extraction cost is defined as the additional penalty accrued in the *Sub-Extraction* conditions (Subjects: A - I; Objects: A - H) that exceeds the combined, predicted costs of complexity and extraction (Subjects: (A - I) - [(A - C) + (A - E)]; Objects: (A - H) - [(A - B) + (A - D)]). This sub-extraction cost is critical for assessing islandhood, as this cost is predicted to be super-additive in island domains, revealing an additional component which exceeds the component costs of complexity and extraction (Sprouse, 2007; Sprouse et al., 2012). Rather than solely assessing the presence of a sub-extraction component for subjects, we define a subject island effect as a reliably greater sub-extraction cost for subjects than for objects. Thus, we do not rely on a subject sub-extraction penalty as the sole measure of islandhood, as in previous studies which have utilized related factorial designs, nor do we interpret the absolute differences between subject and object sub-extraction conditions in isolation.

As discussed in Stigliano et al. (2025), factorial designs for subject islands have varied between comparisons of full extraction and sub-extraction from subject and object positions (e.g. Sprouse et al., 2012: Experiment 1), and comparison of full extraction and sub-extraction from matrix subject and embedded subject positions (e.g. Sprouse et al., 2012: Experiment 2). These two designs implicate distinct components which degrade acceptability. The former aims to account for extraction asymmetries between subjects and objects, and the asymmetry between full extraction and sub-extraction. The latter aims to isolate the cost of a complex versus simple DP, and the asymmetry between matrix and embedded extraction. The present study builds upon these previous designs to incorporate these independent sources of unacceptability in one design. Note, however, that the present study does not compare matrix and embedded extractions, and rather, compares baseline conditions without extraction to conditions with full and sub-extraction. Furthermore, the manipulations in the present design uniquely allow for the calculation of a sub-extraction penalty in both subject and object positions, and the comparison of these two costs across constructions. This comparison of subject and object sub-extraction penalties is our critical measure of a subject island effect, reflecting the residual difference between the sub-extraction cost from a (purported) island domain (i.e. subjects) and the sub-extraction cost from a closely related non-island domain (i.e. objects) which may arise independently.

All things held constant, we anticipate that any sentence which has a complex DP and involves a filler-gap dependency, will receive lower acceptability ratings than its baseline declarative, reflecting the additive costs of DP complexity and of a filler-gap dependency. The cost of sub-extraction*,* therefore, is any additional degradation in acceptability of a given sentence that surpasses the predicted cost of DP complexity and a filler-gap dependency. This value, which follows a similar logic as the calculation of an interaction between the experimental factors in a regression model, is referred to as the DD score (or 'difference of differences' score) in line with the terminology of Sprouse (2007) and Sprouse et al. (2012). The DD score allows us to represent a sub-extraction cost for both subjects and objects, respectively, and to compare the two. This design is particularly suited for abstracting away from constructional differences, as the different constructions we target (WHQ, RC, and TOP) may exhibit independent differences in relative costs of extraction and complexity. For example, the subject advantage is mostly discussed in the case of RCs. If it indeed affects RCs to a greater extent, it is crucial to isolate it in order to better understand observed differences between sub-extraction across constructions, and whether they reflect a contrast in the subject island effect, as Abeillé et al. (2020) suggest, or instead stem from unrelated variation across the construction types.

To preview the results of our experiments, we find that, not only is there a consistently higher DD score for subjects compared to objects across constructions, but that the contrast in DD scores between subject and object positions is stable and nearly constant across constructions, while the absolute ratings of subject and object sub-extraction vary substantially. Thus, despite inherent differences in the acceptability judgments across the three constructions, we observe an invariant difference between subject and object sub-extraction costs, reflecting a subject island effect for all three constructions. Sections 4-6 present the results of three experiments with the super-additive design outlined above, across three construction types (WHQ, RC, and TOP), each with the different IS signatures laid out in Section 2. Section 7 presents the cross-constructional analysis of full and sub-extraction penalties from subject and object positions.

Abeillé et al. (2020) investigated sub-extraction with both pied-piping (fronting an entire prepositional phrase) and p-stranding (fronting a DP and stranding a preposition at the gap site within the relevant extraction domain). Only the pied-piping conditions exhibited the contrast in subject island effects between WHQ and RC dependencies predicted by the FBC, while the p-stranding conditions displayed a consistent subject island effect across WHQs and RCs. In defense of their proposal, Abeillé et al. (2020) claim that pied-piping alleviates an independent parsing difficulty caused by p-stranding inside a subject, and therefore ratings collected under pied-piping are more closely tied to grammaticality. This claim relates to work that has suggested that (subject-internal) gaps, if possible at all, are more acceptable when predictable and plausible, and pied-piping may be advantageous in reducing the set of potential gap positions (Chaves 2013, Chaves & Dery 2019).

The claim that p-stranding out of subjects is rated poorly due to a processing difficulty, and not a subject island effect, attempts to set aside "p-stranding out of subjects" as an exceptional case, despite these being the precise instances that are identified as subject islands. The claim that p-stranding is particularly difficult to process is not well-supported, and overlooks some properties of p-stranding that can in fact improve interpretability, such as the stranded preposition providing an unambiguous cue to the position of the gap, which is crucial for the resolution of any filler-gap dependency. Previous experimental studies have aimed to carefully control for this confound of gap position identification, which can independently impact acceptability ratings (Sprouse et al., 2012; Sprouse et al., 2016).

We argue that p-stranding is the most appropriate environment to test in English. We note additionally that p-stranding in English occurs at a much higher frequency than pied-piping, and belongs to a colloquial register (Enzinna, 2013; Gries, 2002; Günther, 2021; Huddleston & Pullum, 2002). Studies have additionally found that, in English, p-stranding is generally preferred over pied-piping in RCs in acceptability judgement tasks (Hoffmann, 2011; Huddleston & Pullum 2002; Trotta, 2000). A summary of our justifications for utilizing p-stranding in the present study is provided in Table 2, and discussed in more detail in Section 8.

| | **Motivations for utilizing p-stranding in this study** |
|---|---|
| | ◆ P-stranding is more common than pied-piping in English, and belongs to a colloquial register (Enzinna, 2013; Gries, 2002; Günther, 2021; Huddleston & Pullum, 2002)<br>◆ P-stranding is generally preferred over pied-piping in acceptability judgement tasks (for English RCs in particular) (Hoffmann, 2011; Huddleston & Pullum 2002; Trotta, 2000)<br>◆ P-stranding ensures an unambiguous gap site, avoiding an attachment ambiguity that can arise for pied-piping (Sprouse et al., 2012; Sprouse et al., 2016)<br>◆ Researchers have argued that pied piping and p-stranding instantiate distinct types of dependencies, and that crucially, pied-piping is not a true instance of extraction (Bianchi |
| | & Chesi, 2014; Jurka, 2010)<br>◆ The FBC (Abeillé et al., 2020; Winckel et al., 2025) does not distinguish between pied-piping and p-strandings, and makes the same predictions for both types of filler-gap dependencies. |

**Table 2**. Summary: motivations for utilizing p-stranding in this study.

## 4. Experiment 1: Wh-questions (WHQ)

Experiment 1 was designed to test whether subjects are islands for WHQ extraction. The cost of subject sub-extraction versus object sub-extraction is calculated using a super-additive design, which factors out the baseline costs of DP complexity and extraction.

Since a fronted *wh*-phrase is focused, subject sub-extraction for the creation of a WHQ is predicted to introduce a clash in IS, where a focused constituent (the *wh*-phrase) is a sub-part of a backgrounded constituent (the subject). Thus, a WHQ targeting an element within a subject is predicted to be unacceptable under both a *constructional IS profile* account, and a syntactic account that does not distinguish between different discourse functions associated with extraction. Thus, Experiment 1 allows us to test the validity of our super-additive design, and see whether it is able to replicate uncontroversial subject island effects.

### 4.1 Methods

#### 4.1.1 Materials

Thirty-six item sets were created, manipulating Position (extraction domain, *subject* or *object*), DP Complexity (*simple* or *complex*), Extraction Type (*no*, *full*, or *sub-extraction*). The *no extraction* conditions were declarative sentences with no filler-gap dependency, the *full extraction* conditions were WHQs targeting the relevant DP, leaving a gap in the argument position, and the *sub-extraction* conditions were WHQs targeting an embedded DP within the relevant DP. A sample itemset for this experiment can be found in Table 1. For access to all materials, including a full list of itemsets and fillers for all experiments presented in this paper, our experiment code, results, and analysis files, visit our OSF data repository.

The critical object and subject appeared in an embedded declarative clause, and the conditions with extraction involved extraction across an embedded clause. The experimental items targeted DP extraction with preposition stranding within the DP. The position of each critical DP was counterbalanced across item sets, such that, for example, 'the investigator of the crime' was used once as a subject (in Table 1) and once as an object, in another item set. This ensured that DP position (object/subject) is not confounded with the internal makeup of the DP itself, as extraction of some DPs may be more natural than others, independent of their position.

In addition, the experimental items balanced the animacy of the four critical nouns: the subject head noun and its complement in complex subject conditions, and the object head noun and its complement in complex object conditions. This ensured that a contrast between subject and object extraction is not specific to some animacy configurations, as, for example, animate fillers are reportedly better as subjects (e.g., Gennari & McDonald, 2008).

Experimental items were assigned to 9 lists in a Latin square design, and combined with 72 filler items. The filler items varied in acceptability and complexity, including: simple declaratives, it-clefts, unlicensed gaps, and polar questions. The filler items were selected to encourage participants to use the full range of the acceptability scale, and to reduce the likelihood that participants would identify WHQs as a distinct or marked phenomenon within the task. To probe for the effectiveness of the filler items in concealing the experimental manipulations, each experiment was first piloted by a sampling of undergraduate students at UCSC, who received compensation in the form of course credit. These pilots revealed that the fillers were sufficient in masking the critical manipulations in the target items and encouraged the use of the full acceptability scale.

### 4.1.2 Participants and procedure

72 self-reported English speakers participated in this experiment. Participants were recruited on Prolific, and were compensated at a rate of $12 per hour.

The experiment was run on the online experiments platform PCIbex (Zehr & Schwarz, 2018). After providing informed consent and answering demographic questions, participants were instructed that they would be rating the acceptability of sentences in English on a 6-point scale, using the number keys on their keyboard. A fully acceptable sentence was defined as "a sentence that sounds like natural, 'grammatical' English that you might imagine a friend or colleague using", while a fully unacceptable sentence was defined as "a sentence that contains a clear error that you don't think any English speaker would make". Participants were encouraged to use their gut instinct when making their judgments.

The experiment began with a practice session, consisting of 4 trials. At each trial a sentence was displayed at the center of the screen, accompanied by a 6-point acceptability scale below it. Participants were not restricted by a time limit to provide their response.

### 4.1.3 Analysis

Results were analyzed using the R statistical computing environment (R Core Team, 2024) and modeled with a series of Bayesian ordinal mixed-effects regressions assuming a cumulative logit link function using the `brms` package (Bürkner, 2021). The model included fixed effects of the experimental factors, DP Complexity, Extraction Type, Position, and their interaction. The model also included random intercepts for both participant and item. Prior to modeling, we excluded the *Complex~Full Extraction* condition from the analysis for the model to estimate the intended comparisons for measuring an island effect. An effect of DP Complexity reflects the complexity cost by comparing ratings of *Simple~No Extraction* conditions to ratings of *Complex~No Extraction* conditions. An effect of Extraction reflects the extraction cost by comparing ratings of *Simple~No Extraction* conditions to *Simple~Full Extraction* conditions. With the exclusion of the *Complex~Full Extraction* condition, the estimation of the interaction of DP Complexity and Extraction involves comparing the difference between the extraction cost with simple DPs (*Simple~No Extraction*, *Simple~Full Extraction*) and the extraction cost with complex DPs (*Complex~No Extraction, Complex~Sub-extraction*). Thus, this interaction term reflects a super-additive sub-extraction cost, or the additional cost of extraction from a complex DP that cannot be attributed to the independent costs of DP complexity or extraction. The three-way interaction effect between DP Complexity, Extraction Type, and Position reflects a difference in the magnitude of this sub-extraction cost between Subject and Object extractions. Experimental factors were modeled using sum contrast coding (-0.5, 0.5), with the subject conditions, complex conditions, and extraction conditions as the positive coefficients. Following Kush et al. (2018), Kush et al. (2019), we report only these interaction effects, as the main effects of the experimental factors and the additional interaction terms are irrelevant to the central questions regarding the islandhood of subjects.

In addition to reporting these interaction terms, we report Differences-in-Differences (DD) scores as another measure of an island effect, calculated on ratings *z*-scored by participant (following the analysis reported in Vincent et al., 2022). This is calculated by first estimating the costs of DP complexity and Extraction. The cost of DP Complexity is calculated as the difference in mean ratings of *Simple~No Extraction* conditions and *Complex~No Extraction* conditions, and the cost of Extraction is calculated as the difference in mean ratings of *Simple~No Extraction* and *Simple ~ Full Extraction* conditions. The DD score corresponding to a sub-extraction cost compares the observed ratings of sub-extraction conditions to the predicted costs of DP Complexity and Extraction, whereby a positive DD score indicates an additional penalty that cannot be attributed to the costs of Complexity or Extraction as observed in the study. Thus, these DD scores reflect identical comparisons to those implemented in the model estimations described above, though in different scales. As discussed by Kush et al. (2018), Sprouse (2007), Sprouse et al. (2012), Vincent et al. (2022), among others, DD scores provide a standardized measure of island effect size, with DD scores closer to zero reflecting the absence of an island effect.

### 4.2 Results

The mean raw ratings from Experiment 1 with WHQs are plotted in Figure 1. This figure also provides the DD scores, calculated from the ratings z-scored by participant. As expected, across both *Object* and *Subject* conditions, *Complex DP* conditions were rated lower than *Simple DP* conditions, and *Full Extraction* conditions were rated lower than *No Extraction* conditions. The *Object* and *Subject* conditions received similar ratings overall, whereby the apparent costs of Complexity and Extraction are nearly identical for Subjects and Objects. In sub-extraction conditions, *Subject~Sub-extraction* conditions received lower ratings than *Object~Sub-extraction* conditions, indicative of the presence of a subject island effect.

Results from the ordinal mixed-effects regression model revealed an interaction effect between DP Complexity and Extraction Type [$\beta = -0.95$, 95%CrI = (-1.26, -0.64), Std. Error = 0.16, $Pr(\beta < 0) = 1.00$], reflecting the presence of a super-additive cost of sub-extraction. We also observed a three-way interaction between DP Complexity, Extraction Type, and Position, indicating a significant difference in the super-additive sub effect between Subjects and Objects [$\beta = -0.94$, 95%CrI = (-1.54, -0.32), Std. Error = 0.31, $Pr(\beta < 0) = 0.99$]. This interaction term indicates a larger sub-extraction cost for Subjects than Objects in WHQs, corroborating the difference in DD scores shown in Figure 1.

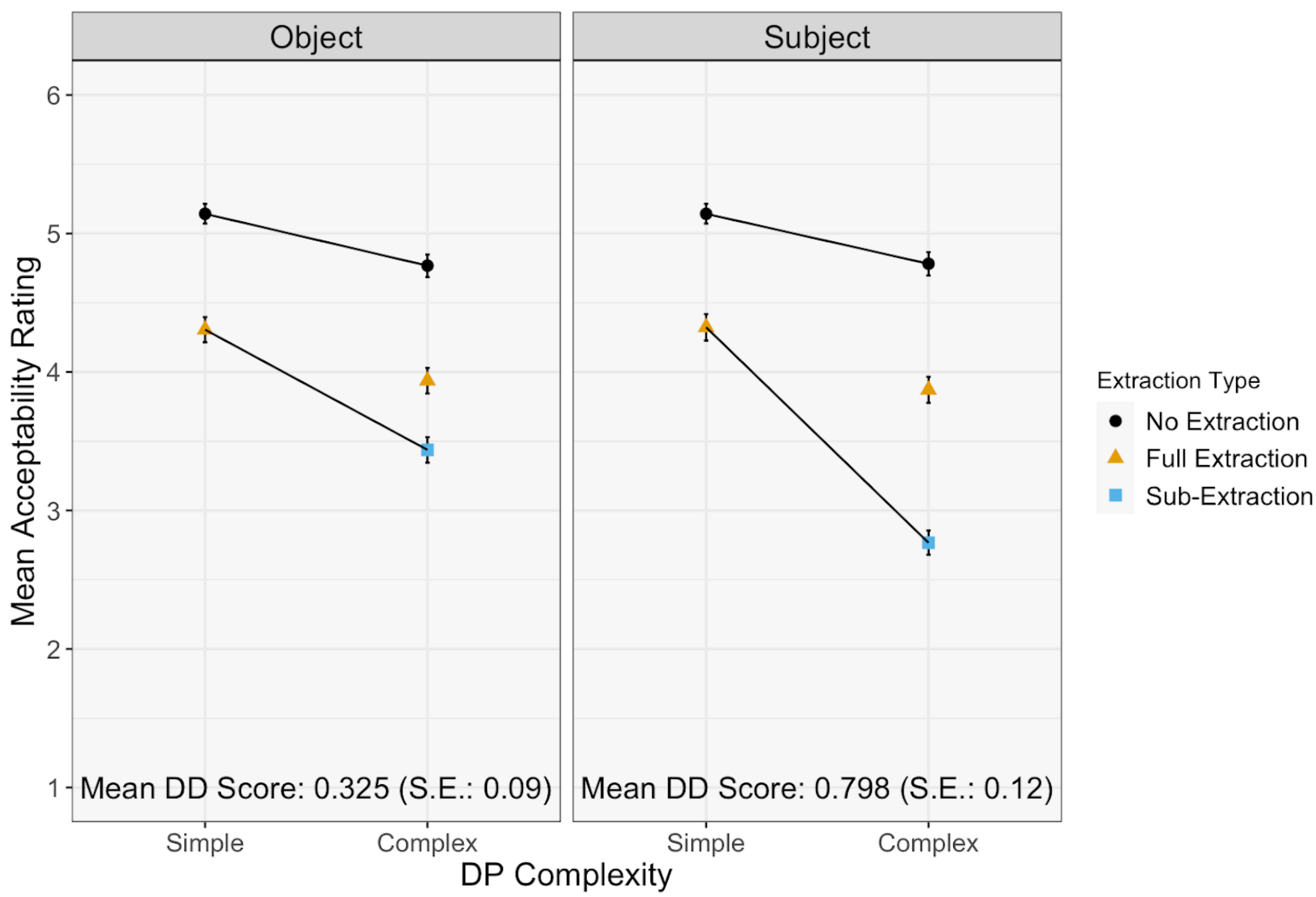

Figure 1: Experiment 1 (WHQ) mean ratings, faceted by Position, arranged in columns by DP Complexity. Error bars represent standard error.

### 4.3 Discussion

The results of Experiment 1 provide evidence for a subject island effect in WHQs. The cost of subject sub-extraction exceeds its predicted cost based on DP-complexity and extraction. That is, the degradation associated with sub-extraction was larger than the combined degradation of a complex DP and of extraction (compared to baseline declarative sentences). While this was true for both subjects and objects, the additional cost of sub-extraction was *larger* for subject extraction compared to object extraction (as evidenced by the three-way interaction, and the contrast in DD scores), suggesting that subject sub-extraction is particularly degraded.

The WHQ results suggest that while subjects and objects do not differ much with respect to the acceptability of full extraction and of hosting a complex DP, which incur some cost in both, the acceptability of subject sub-extraction is disproportionately worse than object sub-extraction. Thus, Experiment 1 neatly replicates the wealth of previous findings that subjects are islands for WHQs (Abeillé et al., 2020, Kush et al., 2018, Sprouse et al., 2012, among others). In the following experiments, we adopt an identical design to investigate subject islandhood with RCs (Experiment 2) and TOPs (Experiment 3).

## 5. Experiment 2: Relative clauses (RCs)

Experiment 2 was designed to test whether subjects are islands for RCs. Like Experiment 1, the cost of subject sub-extraction versus object sub-extraction is calculated using a super-additive design, which factors out the baseline costs of DP-complexity and extraction in each position.

Unlike WHQs, the filler in RCs is not focused. Thus, under a *constructional IS profile* account, RCs should not display a subject island effect, since subject sub-extraction does not introduce an IS clash: both the subject and the filler (the head of the RC) are positions that normally introduce backgrounded or presupposed arguments. However, a syntactic account that attributes island effects to the grammatical operation of extraction would predict RCs to produce a subject island effect, no different from WHQs.

### 5.1 Methods

#### 5.1.1 Materials

The experimental items in Experiment 2 were based on the items in Experiment 1, using the same DPs and embedded verbs, but with RCs instead of matrix WHQs. Thirty-six item sets were created, manipulating Position (extraction domain, *subject* or *object*), DP Complexity (*simple* or *complex*), Extraction Type (*no*, *full*, or *sub-extraction*). A sample itemset is given in Table 3 below.

| | |
|---|---|
| *No Extraction* | |
| Simple | I noticed that [ Stephanie explained the investigator had already questioned the driver ]. |
| Complex object | I noticed that [ Stephanie explained the investigator had already questioned the driver of the car ]. |
| Complex subject | I noticed that [ Stephanie explained the investigator of the crime had already questioned the driver ]. |
| *Simple DP, Full Extraction* | |
| Object | I noticed [ the driver that Stephanie explained the investigator had already questioned _ ]. |
| Subject | I noticed [ the investigator that Stephanie explained _ had already questioned the driver ]. |
| *Complex DP, Full Extraction* | |
| Object | I noticed [ the driver of the car that Stephanie explained the investigator had already questioned _ ]. |
| Subject | I noticed [ the investigator of the crime that Stephanie explained _ had already questioned the driver ]. |
| *Complex DP, Sub-Extraction* | |
| Object | I noticed [ the car that Stephanie explained the investigator had already questioned the driver of _ ]. |
| Subject | I noticed [ the crime that Stephanie explained the investigator of _ had already questioned the driver ]. |

**Table 3**. Example set from Experiment 2.

Since RCs do not correspond to full sentences, but rather instead create DPs, the materials in Experiment 2 included an additional predicate. In particular, the RC head in the extraction conditions was the object of a matrix verb. In order to balance the overall complexity of the extraction and non-extraction conditions, we selected matrix verbs that could embed both an argument and a clause. This allowed us to use the same matrix verbs across conditions, such that the no-extraction declarative was also embedded.

Experimental items were assigned to 9 lists in a latin square design, and combined with 72 filler items. The filler items varied in acceptability, complexity and whether they involve extraction. The filler items were selected to encourage participants to use the full range of the acceptability scale and to minimize the likelihood that the RCs in our target items drew any particular attention.

### 5.1.2 Participants, procedure, and analysis

72 self-reported English speakers participated in this experiment. Participants were recruited on Prolific, and were compensated at a rate of $12 per hour. The procedure and analysis of Experiment 2 were identical to that of Experiment 1.

## 5.2 Results

The mean raw ratings from Experiment 2 with RC extraction are plotted in Figure 2, as well as the DD scores calculated from the ratings z-scored by participant. Across both *Subject* and *Object* conditions, *Complex DP* conditions received lower ratings than *Simple DP* conditions. *Full Extraction* conditions received lower ratings than *No Extraction* conditions, and this contrast was larger for object extractions. Crucially, the super-additive cost of sub-extraction was larger in subjects compared to objects, despite the overall large baseline cost of extraction observed in this experiment.

Results from the ordinal mixed-effects regression model revealed an interaction effect between DP Complexity and Extraction Type [$\beta = -0.67$, 95%CrI = (-0.96, -0.37), Std. Error = 0.15, $Pr(\beta < 0) = 1.00$], indicating the presence of a significant super-additive cost of sub-extraction. We crucially observed a three-way interaction between DP Complexity, Extraction Type, and Position, indicating a significant difference in the super-additive sub-extraction cost between Subjects and Objects [$\beta = -0.58$, 95%CrI = (-1.17, 0), Std. Error = 0.30, $Pr(\beta < 0) = 0.98$]. This interaction term indicates a larger sub-extraction penalty for Subjects than Objects with RC extraction, corroborating the difference in DD scores shown in Figure 2.

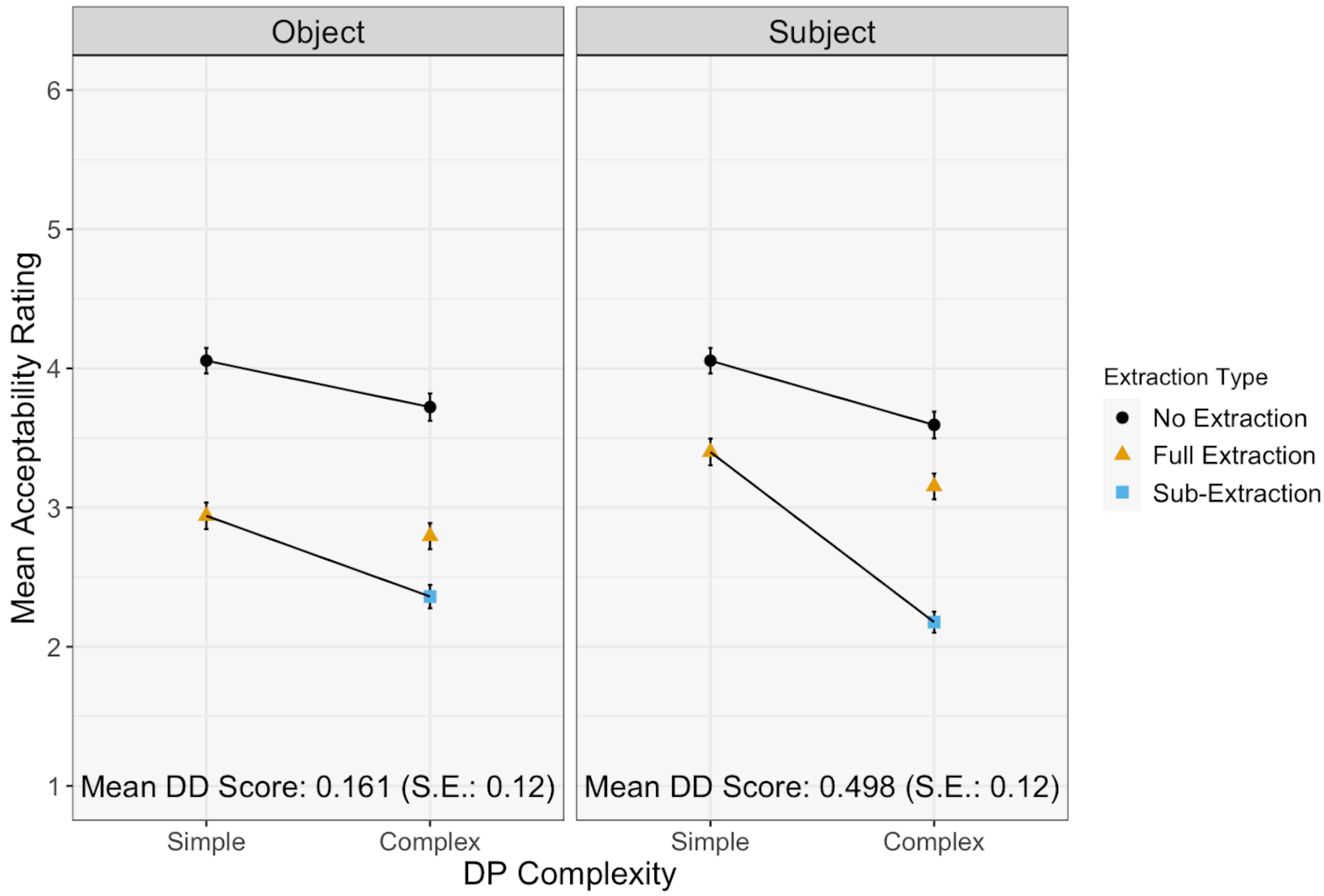

Figure 2: Experiment 2 (RC) mean ratings, faceted by Position, arranged in columns by DP Complexity. Error bars represent standard error.

**5.3 Discussion**

The results of Experiment 2 provide evidence for a subject island effect in RCs. Much like in WHQs, we found evidence that the cost of subject sub-extraction exceeds its predicted cost based on DP-complexity and extraction. That is, the degradation associated with sub-extraction was larger than the combined degradation of a complex DP and of extraction, compared to baseline declarative sentences. While this was true for both subjects and objects, the additional cost of sub-extraction was again *larger* for subject extraction compared to object extraction, as evidenced by the three-way interaction, and the contrast in DD scores, suggesting that subject sub-extraction is particularly degraded. The combined results of Experiments 1 and 2 conceptually replicate existing cross-linguitic findings that subjects are islands for both WHQ and RC dependencies (Kobzeva et al., 2022; Sprouse et al., 2016).

This result is incompatible with the predictions of *constructional IS profile* accounts for subject islandhood, in particular, the FBC (Abeillé et al., 2020). Since sub-extraction from a subject for the purpose of creating an RC does not introduce a clash in IS, as both subjects and RC fillers are given, the *constructional IS profile* accounts predict no subject island effect to arise in RCs. Instead, this finding supports the view of islands as syntactic conditions, sensitive to constraints on abstract computation rather than discourse function.

While in Experiment 1, targeting WHQs, the mean acceptability rating of subject sub-extraction seems lower than object sub-extraction, this was not the case for the RCs in Experiment 2. However, for both construction types, the super-additive cost of sub-extraction was larger in subjects compared to objects, indicating a subject island effect. This difference between the costs of subject versus object sub-extraction in RCs reflects the fact that (full) extraction in RCs is worse when targeting objects compared to subjects (see Lau & Tanaka 2021 for a recent overview). This contrast, which is not present in WHQs, is predicted to independently lower the ratings of object sub-extraction, obscuring a direct comparison between subject and object sub-extraction in RCs. Only when taking into account this baseline contrast between subject and object extraction in RCs, using our super-additive design, were we able to detect a subject island effect in this construction. Notably, this effect may go undetected in less granular designs.

This underscores the importance of controlling for additional factors that influence the acceptability of sub-extraction across positions. Only when taking into account how subjects and objects might be affected differently by the grammatical manipulations that make up sub-extraction (complexity and extraction), can island effects be appropriately isolated. We return to this point in section 8, where we discuss Abeille et al. (2020)'s claim that RCs do not induce a subject island effect.

## 6. Experiment 3: Topicalization (TOP)

Experiment 3 was designed to test whether subjects are islands for TOPs. Like the previous experiments, the cost of subject sub-extraction versus object sub-extraction is calculated using a super-additive design, which factors out the baseline costs of DP-complexity and extraction.

Much like RCs, but unlike WHQs, a topicalized element is not focused. Thus, under a *constructional IS profile* account, TOP should not display a subject island effect, since both the subject and the extracted topic are positions that normally introduce backgrounded or presupposed elements. However, a syntactic account which attributes island effects to the grammatical operation of extraction would predict TOPs to display the same subject island effect found in other construction types.

In addition, since a topicalized constituent is normally interpreted as given, avoidance of IS clashes within a constituent would predict topicalization out of an *object* to be degraded. That is, if we generalize the ban on focusing a sub-part of a backgrounded constituent, at the heart of *constructional IS profile* accounts, to a ban on any intra-constituent IS clash, we predict *objects* to be islands in TOPs. This is because object sub-extraction would result in an extracted element which is given (the topic), but is a sub-part of a constituent often associated with new information (the object).

### 6.1 Methods

#### 6.1.1 Materials

The experimental items in Experiment 3 were based on the items in Experiment 1, using the same DPs and embedded verbs, but with TOP constructions instead of matrix questions. Thirty-six item sets were created, manipulating Position (extraction domain, *subject* or *object*), DP Complexity (*simple* or *complex*), Extraction Type (*no*, *full*, or *sub-extraction*). A sample itemset is given in Table 4 below.

| | |
|---|---|
| *No Extraction* | |
| Simple | Stephanie explained the investigator had already questioned the driver. |
| Complex object | Stephanie explained the investigator had already questioned the driver of the car. |
| Complex subject | Stephanie explained the investigator of the crime had already questioned the driver. |
| *Simple DP, Full Extraction* | |
| Object | That driver, Stephanie explained the investigator had already questioned _ . |
| Subject | That investigator, Stephanie explained _ had already questioned the driver. |
| *Complex DP, Full Extraction* | |
| Object | That driver of the car, Stephanie explained the investigator had already questioned _ . |
| Subject | That investigator of the crime, Stephanie explained _ had already questioned the driver. |
| *Complex DP, Sub-Extraction* | |
| Object | That car, Stephanie explained the investigator _ had already questioned the driver of _ . |
| Subject | That crime, Stephanie explained the investigator of _ had already questioned the driver. |

**Table 4**. Example set from Experiment 3.

Experimental items were assigned to 9 lists in a Latin square design, and combined with 72 filler items. The filler items varied in acceptability, complexity and whether they involve extraction. For example, this set of filler items included simple declarative sentences with demonstrative subjects and declaratives with sentence-initial temporal adverbials to mirror the left-dislocated topics in the target items. The filler items were selected to encourage participants to use the full range of the acceptability scale and to minimize the likelihood of our target items drawing particular attention.

#### 6.1.2 Participants, procedure, and analysis.

72 self-reported English speakers participated in this experiment. Participants were recruited on Prolific, and were compensated at a rate of $12 per hour. The procedure and analysis for Experiment 3 was identical to that of Experiments 1 and 2.

### 6.2 Results

The mean raw ratings from Experiment 3 with TOP constructions are plotted in Figure 3, along with mean DD scores calculated from the ratings z-scored by participant. While *Complex DP* conditions received lower ratings than *Simple DP* conditions for items with no extraction, in *Full Extraction* conditions, this difference is substantially attenuated for object extraction, but not subject extraction. As in Experiment 2 with RCs, *Full Extraction* conditions received lower ratings than *No Extraction* conditions, while sentences with (full) extractions of subjects received higher ratings than those with object extractions. Sub-extractions from subjects and objects received (nearly identical) low ratings, in contrast to the difference between ratings of full extractions of subjects and objects observed in Figure 3.

Results from the ordinal mixed-effects regression model revealed a three-way interaction between DP Complexity, Extraction Type, and Position, indicating a significant difference in the super-additive sub-extraction cost between Subjects and Objects [$\beta = -1.24$, 95%CrI = (-1.90, -0.59), Std. Error = 0.33, $Pr(\beta < 0) = 0.99$]. This interaction term indicates a larger sub-extraction penalty for Subjects than Objects with RC extraction, in accord with the difference in DD scores shown in Figure 3. We did not find an interaction effect between DP Complexity and Extraction Type [$\beta = -0.01$ , CrI = ( -0.33, 0.32), Std. Error = 0.16, $Pr(\beta < 0) = 0.52$].

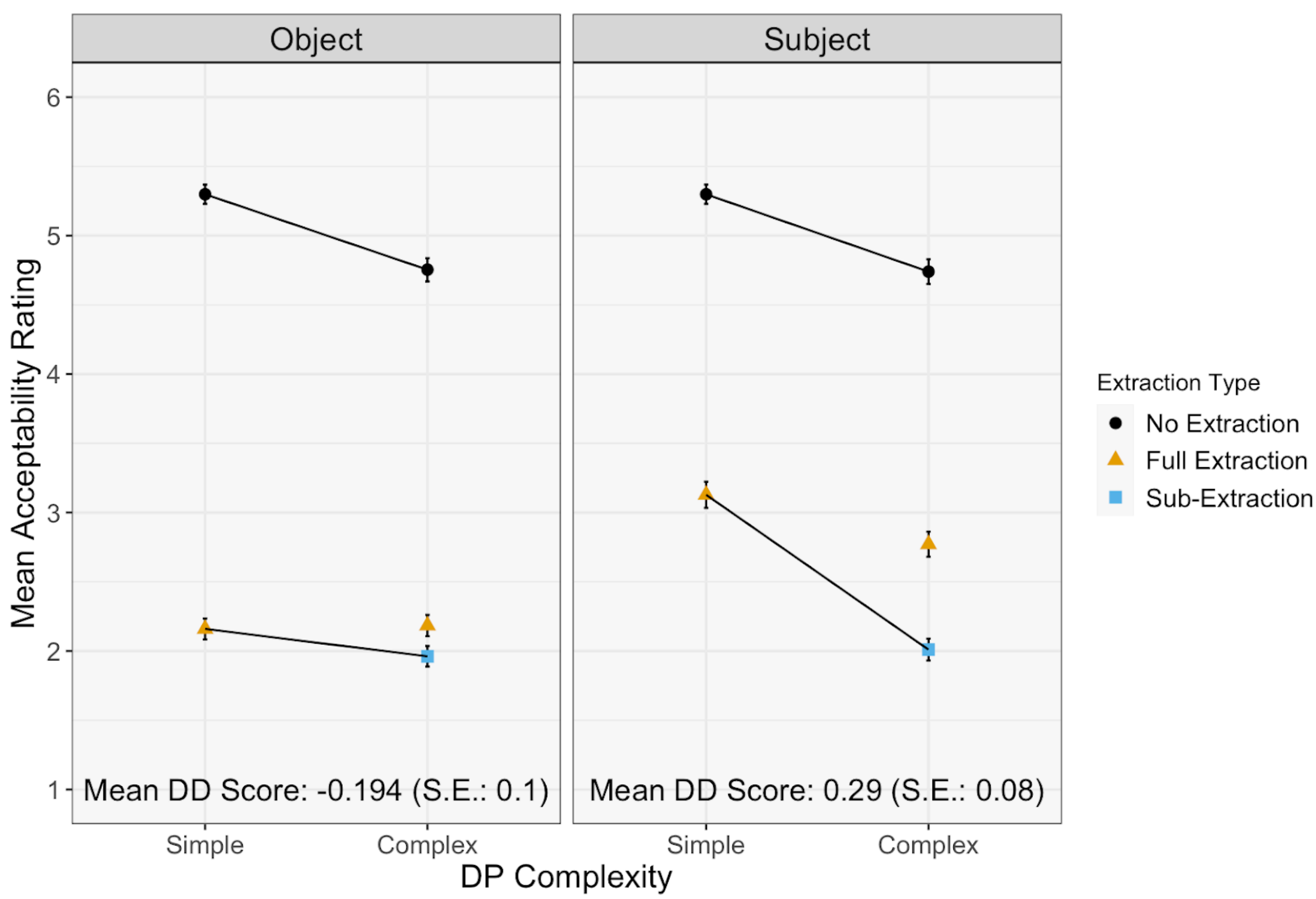

Figure 3: Experiment 3 (TOP) mean ratings, faceted by Position, arranged in columns by DP Complexity. Error bars represent standard error.

### 6.3 Discussion

The results of Experiment 3 provide evidence for a subject island effect in TOP, much like in WHQs and RCs. We found evidence that for TOP constructions, the cost of subject sub-extraction exceeds its predicted cost based on DP-complexity and extraction, while this was not the case for object sub-extraction (as evidenced by the three-way interaction, and the contrast in DD scores). This provides further evidence against *constructional IS profile* accounts, which characterize the subject island effect as a restriction on focusing a sub-part of a backgrounded constituent, as the filler in TOP constructions, like RCs, is backgrounded and not focused. The subject island effect observed in Experiment 3 is consonant with that found by Kush et al. (2019) for TOP dependencies in Norwegian, and in tandem with the results of Experiments 1 and 2, contribute to the cross-linguistic finding of robust subject island effects across dependencies which differ in their discourse function (Kobzeva et al., 2022; Kush et al., 2018, 2019; Sprouse et al., 2016).

As in RCs, we observed no difference between the absolute ratings of object and subject sub-extraction in TOP. However, only subjects displayed a positive super-additive cost of sub-extraction, when taking into account the different costs of complexity and TOP extraction in subjects and objects. In particular, like RCs, full TOP extraction of objects is degraded compared to full extraction of subjects. This again demonstrates the importance of using a super-additive design for the diagnosis of island effects, as the absolute acceptability rating of an island violating construction may appear to be identical to that of a non-island violating construction. When controlling for the component costs that contribute to these degradations in acceptability, however, we find that sub-extraction from subjects is rated lower than these predicted baseline measures, indicating the presence of an additional island cost that cannot be attributed to these measures. This profile is entirely absent for object conditions, where we found evidence suggesting a subadditive cost of sub-extraction.

## 7. Comparing constructions

In each experiment, we found an interaction of DP Complexity, Extraction Type, and Position, which corresponds to a greater super-additive cost for subject sub-extraction than object sub-extraction. The absolute cost for subject sub-extraction (DD-score), however, varied across the three constructions. To better understand this variation, we conducted a cross-constructional analysis. We fit a set of ordinal mixed-effects regression models assuming a cumulative logit link function, using the `brms` package (Bürkner, 2021) to conduct pairwise comparisons between each condition to the baseline *Simple~No Extraction* condition for each experiment. We use coefficients from the ordinal regression, which express differences in terms of standard variates, to standardize and directly compare ratings across the three experiments. We take the posterior draws from our regression models using the `tidybayes` package (Kay, 2024), and use the logic of our DD score calculations to express the costs of sub-extraction and full extraction for both subjects and objects as posterior distributions with 95% HPDI. This provides an additional estimate of the unacceptability of sub-extraction compared to full complex DP extraction across constructions (WHQs, RCs and TOPs).

The comparison of the *Complex~No Extraction* condition to the baseline provides an estimate of the cost of complexity, and the comparison of the *Simple~Full Extraction* condition to the baseline provides an estimate of the cost of extraction. DD scores for both sub-extraction and full extraction costs were calculated by subtracting the sum of the complexity and extraction costs from the model estimates of the *Complex~Sub-extraction* and *Complex~Full Extraction* conditions, respectively. The full extraction costs are predicted to be centered around 0, as the *Complex~Full Extraction* conditions are expected to reflect the additive costs of complexity and extraction. Sub-extraction costs are predicted to exceed 0, as we expect there to be an additional cost of sub-extraction that is not attributed to the independent costs of complexity and extraction. Moreover, this sub-extraction cost is expected to be largest for subjects, indicating the presence of an island violation.

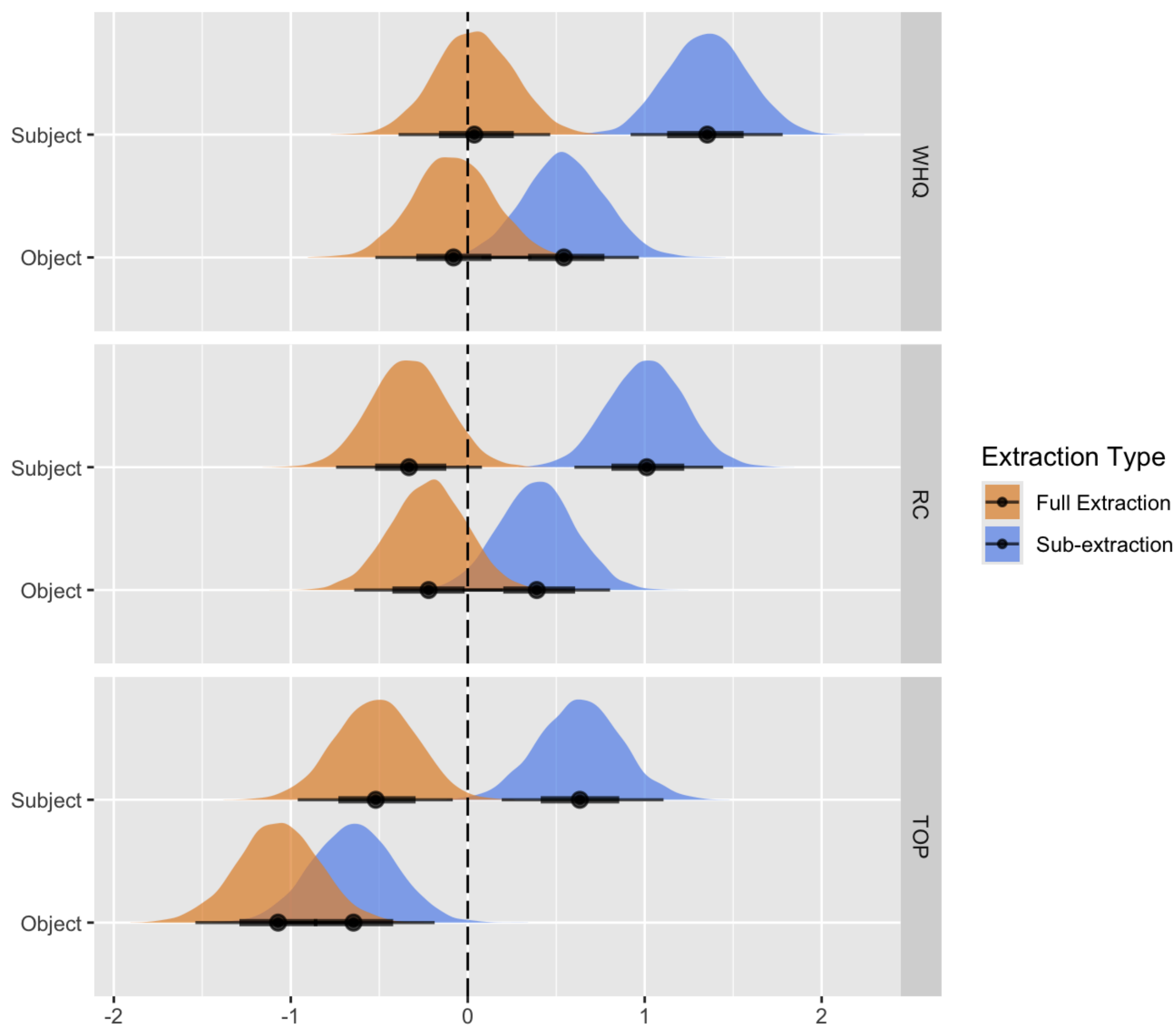

Figure 4: Posterior distributions of the standardized extraction costs by position, faceted by construction.

Figure 4 plots the posterior distributions of our DD scores for each of the positions (Subject, Object) and extraction types (Full, Sub) in each experiment. We find a strikingly consistent pattern across experiments. For subjects in each experiment, the distributions of estimated sub-extraction costs substantially exceed the distributions of estimated full-extraction costs. For objects, however, these distributions show considerable overlap, indicating a smaller difference between sub-extraction and full extraction costs. Additionally, while the distribution of full extraction costs for subjects and objects have similar means and 95% HPDIs, the distribution of sub-extraction costs for subjects substantially exceeds that of objects. The consistent difference between subject and object DDs across constructions is what we identify as a subject island effect.

Despite these similarities for each experiment, there are apparent differences in the absolute size of the sub-extraction cost between the experiments, with the largest sub-extraction DD-score for WHQs, a smaller DD-score with RCs, and the smallest DD-score in TOP constructions. That is, when considering the isolated penalty of sub-extraction from subjects, we observe a cline across the construction types, superficially consistent with the characterization of these constructions according to the *constructional IS profile* approach(es): the largest penalty is observed for WHQs, which the FBC predicts to consistently engender an IS clash, and the smallest penalty is observed for TOPs, which the FBC predicts to consistently avoid an IS clash. While the FBC does not predict any penalty for RC extraction nor TOP, the present results are nominally consistent with a view in which different constructions modulate *the degree of* an island effect, rather than the presence/absence of an island effect, as in the recent formalizations of the FBC.

We suggest, instead, that the present results militate against considering the subject sub-extraction cost in isolation. Figure 5 plots the difference between the sub-extraction and full extraction costs for subjects and objects across the three constructions, and reveals a stable and invariant pattern across the constructions. For subjects, the additional cost of sub-extraction is 1.32 (95% HPDI: 1.02, 1.61) in WHQs, 1.34 (95% HPDI: 1.04, 1.64) in RCs, and 1.15 (95% HPDI: .85, 1.45) in TOP constructions. This indicates that despite multiple baseline differences between these constructions, subject sub-extraction is consistently degraded compared to subject full extraction, and to a similar extent across constructions. Similarly, this difference between extraction costs is consistently larger for subjects than objects. Thus, with the appropriate comparisons for each construction, we find that (i) subject sub-extraction is less acceptable than subject full extraction, and that (ii) this difference in extraction costs is greater for subjects than objects. Although the isolated measures suggest apparent differences in 'island effect' size across the three constructions, we suggest that the comparisons of relevant conditions in our factorial design reveal a profile which unambiguously identifies subjects as islands independent of the IS profile of each construction type.

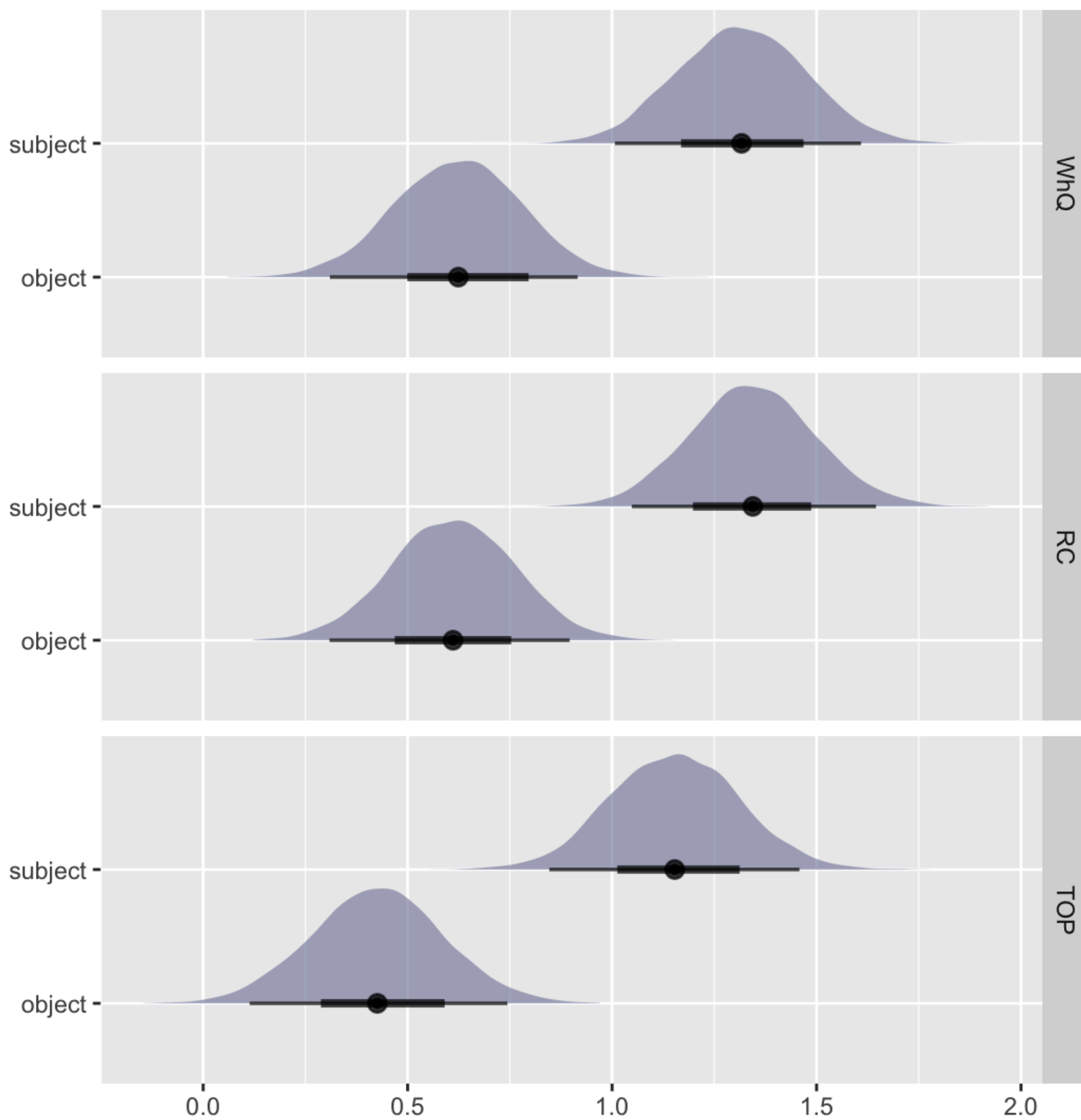

Figure 5: Posterior distributions of the differences between sub-extraction and full extraction by position, faceted by construction.

## 8. General Discussion

Across all three experiments, we found evidence for subject island effects in TOP constructions, WHQs, and RCs: the cost of sub-extraction from subjects was reliably greater than the cost of sub-extraction from objects, when taking into account the costs of DP complexity and extraction in each position. This indicates the presence of an additional penalty associated with sub-extraction from subjects that is not predicted by the independent costs of DP complexity or extraction. Sub-extraction from objects also exceeded the predicted cost, but to a significantly smaller extent than subjects, providing evidence that the subject position is differentially sensitive to sub-extraction, more so than the object position. We conclude that subjects are islands across TOP constructions, WHQs, and RCs, despite IS differences between these constructions.

Our results show that the ban on subject sub-extraction cannot be solely attributed to construction-specific discourse-based preferences. Instead, it is best attributed to the grammatical dependency shared across the different constructions: movement. A close look comparing across the three experiments reveals a stable difference in the DD scores of subjects vs. objects irrespective of construction type, which we take to indicate stability of a singular underlying constraint that regulates the grammatical operation of movement. This analysis contributes an alternative construal of what it means to be a "subject island," which is relative, rather than absolute: here, a subject island effect is measured through the relative degradation of sub-extraction from a DP in a preverbal subject position, compared to the same sub-extraction operation from a DP in a different position — in this case, the object position. Moreover, the experimental manipulation and analyses presented here further highlight the significance of using a factorial design in the study of islands, especially when the goal is to compare across constructions, which likely vary in their sensitivity to extraction from distinct syntactic positions.

We note that our results only speak to the FBC, and do not contradict *direct backgroundedness* approaches to islands (Cuneo & Goldberg, 2023; Goldberg, 2006, 2013; Goldberg et al., 2024). That is because we only manipulated the potential clash between the filler and the gap site, and did not directly manipulate the backgroundedness of the two domains. If subjects were systematically more backgrounded than objects across each of the three tested constructions, then direct backgroundedness could capture our findings.

### 8.1. Contrasts and convergences with previous findings

While we report a subject island effect across dependency types (WHQs, RCs and TOPs), Abeillé et al. (2020) suggest that RCs do not produce a subject island effect, and this motivates their claim that subject island effects are due to a clash in IS. Below, we discuss Abeillé et al. (2020)'s methodology and results, and we argue that their rejection of a subject island effect in RCs is not well-supported.

Abeillé et al. (2020) tested the acceptability of subject and object sub-extraction in WHQs and non-restrictive RCs, in English and French. When sub-extraction of a DP in English was tested in the context of *p-stranding*, subject sub-extraction was rated worse than object sub-extraction in both RCs (18) and WHQs (19). This contrast is consistent with a subject island effect in RCs, as predicted by syntactic accounts of subject islandhood, and as replicated in our study. It is incompatible with the FBC, according to which subjects are only islands for focusing.

(18) The dealer sold a sportscar, which [the color of __] delighted the baseball player because of its surprising luminance.

(19) Which sportscar did [the color of __] delight the baseball player because of its surprising luminance? Abeillé et al. (2020), p. 8 & p. 10

Nonetheless, Abeillé et al. claim that the unacceptability of p-stranding sub-extraction out of a subject is not a subject island effect. According to them, this unacceptability is due to a processing difficulty which arises specifically with DP sub-extraction out of a subject, independent of construction type. In particular, they suggest that sub-extraction out of a subject is rare and hence difficult to process, and that this processing difficulty is exacerbated when the extracted element is a p-stranded DP, compared to a PP. This is due to the restricted distribution of lexical items that select for a PP versus a DP, making a subject internal DP-gap particularly surprising for the parser. We note, however, that for a given lexical nominal head, there are no selectional differences between a DP and its containing PP.

We question Abeillé et al's dismissal of the subject island effect in RCs, and their alternative explanation. First, the argument that subject sub-extraction is unacceptable because it is rare runs the risk of circularity, as it doesn't address the underlying reasons for its rarity. This is precisely the puzzle of subject islands: why are these short dependencies, with a clear intended meaning, rarely attested and judged as unacceptable?

Second, Abeillé et al's processing-based dismissal of island effects with p-stranding rests on the non-trivial assumption that filler-gap dependencies are more difficult when the filler is a DP, as it is with p-stranding, compared to a PP. However, contrary to this assumption, and despite the prescriptivist rules against p-stranding, studies find that p-stranding (DP extraction) is generally preferred over pied-piping (PP extraction) in RCs, both in their relative distribution, and in acceptability judgement tasks (Hoffmann, 2011; Huddleston & Pullum, 2003; Trotta, 2000). Gries (2002) analyzed instances of pied-piping and p-stranding in the British National Corpus, and found that p-stranding makes up a 100% of the spoken utterances, while in written text the distribution of p-stranding (49.39%) and pied-piping (50.61%) is roughly equal. Moreover, RT studies do not support a greater processing difficulty in p-stranding compared to pied-piping (Enzinna, 2013; Günther, 2021). While it is true that PP extraction offers the advantage of a more restricted set of gap positions (the subject or direct object themselves cannot be a PP gap), DP extraction with p-stranding arguably introduces processing advantages of its own. In particular, the gap position may be more readily identified, as the host DP is ungrammatical without the empty argument (*the color of*). When processing a dependency headed by a PP (*of which*), on the other hand, it is not immediately evident that the subject hosts the gap (*the color*), and comprehenders may assume that the extracted PP modifies the verb or another DP in the sentence. These considerations motivated Sprouse et al. (2016)'s factorial design for testing subject islands in Italian[3], in which the gap location of the fronted PP is not unambiguously signaled. To mitigate against a parsing strategy that avoids positing a subject-internal gap or permits an ambiguity in attachment site, their stimuli included a PP adjunct for both subject and object DPs, thus reducing this potential ambiguity regarding the location of the gap.

Jurka (2010) reports a more direct comparison of sub-extraction from subjects and objects in English using both pied-piping and p-stranding in embedded WHQ dependencies. The results of an acceptability judgment study indicated that sub-extraction from subjects was reliably less acceptable than sub-extraction from objects for both pied-piping and p-stranding, indicating that both DP and PP extraction can give rise to an island profile, while the effect with p-stranding was significantly larger than that with pied-piping. Subsequent analysis of the distribution of ratings across conditions revealed that participants consistently dispreferred pied-piping from both subject and object positions. Jurka (2010) interprets these patterns to reflect that pied-piping is a qualitatively distinct dependency from p-stranding, in that it does not involve movement, but is perhaps more akin to a base-generated hanging topic, rather than a 'genuine' filler-gap dependency. Related proposals are put forward by Bianchi & Chesi (2014,

[3] Sprouse et al. (2016) adopted this design for testing subject islands in English (using p-stranding) and Italian, and found that while sub-extraction from subjects exhibited an island profile for both WHQ and RC dependencies in English, subjects were only found to exhibit an island profile with WHQ dependencies in Italian. While this pattern is compatible with the FBC, Sprouse et al. (2016) provide extensive discussion arguing that this cross-linguistic and cross-dependency variation can be understood within various syntactic approaches to islandhood as well (e.g. Chomsky, 1986; Chomsky, 2008; Rizzi, 1982; Rizzi, 2004; Stepanov, 2007).

2015), who argue that PP and DP extraction involve somewhat different dependencies which are interpreted in subtly different ways (in English). Although a full exploration of these arguments is beyond the scope of the present work, these findings provide additional reason for investigating sub-extraction from island and non-island domains with p-stranding in English. Following this work, evidence based on pied-piping in English, as in other languages, may be less than optimal, unless it can be shown that these languages provide unambiguous evidence for movement in pied-piping derivations.

Lastly, not all cases of p-stranding within a subject are considered ungrammatical, contrary to the expectations raised by Abeillé et al. (2020)'s processing based account. In particular, a gap within a subject becomes acceptable when it is accompanied by an additional gap in the clause; this is called a "parasitic gap" (PG) construction (Culicover, 2001; Engdahl, 1983). Phillips (2006) reports experimental data showing that the acceptability of subject sub-extraction (20a) greatly improves when the sentence involves an additional gap (20b). This PG construction is even rated as acceptable as its equivalent with no subject internal gap (20c).

(20) a. The outspoken environmentalist worked to investigate what [the local campaign to preserve _] had harmed the annual migration.

b. The outspoken environmentalist worked to investigate what [the local campaign to preserve _] had harmed _.

c. The outspoken environmentalist worked to investigate what the local campaign to preserve the important inhabitants had harmed _.

While the acceptability of a PG with p-stranding has not yet been experimentally examined, we speculate that the same improvement in acceptability would be observed for subject sub-extraction with p-stranding (21a) when accompanied by a matrix clause gap (21b).

However, if the unacceptability of p-stranding within a RC subject were actually the result of an unrelated processing difficulty, that difficulty should not diminish in a PG construction. If anything, it should get worse— the additional gap arguably complicates the dependency, and since it appears later, it is unclear how it could make the subject-internal gap easier to identify or less surprising (Phillips, 2006; Phillips, 2013; although see Chaves, 2013 for a processing-based proposal for PGs). Without additional stipulations, Abeillé et al.'s account predicts DP sub-extraction from a subject to be unacceptable even in a PG construction, unlike other island violations.

(21) a. The dealer sold a sportscar, which [the attempt to take care of _] ultimately damaged his reputation.

b. The dealer sold a sportscar, which [the attempt to take care of _] ultimately damaged _.

We conclude that the unacceptability of DP sub-extraction from a RC subject, reported in Abeillé et al. and in the current study using a factorial design, cannot be easily explained away as a processing difficulty. In order to develop an alternative account for the RC subject island effect that is based on difficulty or surprisal specific to p-stranding in a subject, one would need to address the high frequency of p-stranding in colloquial English, the online data showing no particular difficulty with p-stranding, and the acceptability of p-stranded parasitic gaps.

Turning now to sub-extraction of a PP in English (i.e., with pied-piping), Abeillé et al. report degraded acceptability for subject sub-extraction in WHQs but not in RCs, based on their design. In RCs, the acceptability ratings for PP sub-extraction from subjects were even slightly higher than PP sub-extraction from objects. This is the finding which forms the basis of the FBC in English. In assessing Abeillé et al.'s conclusion that subjects are not islands for relativizing a PP, it is worth highlighting several details about their design and analysis. First, they compare PP extraction out of a subject and an object directly, and find that subject sub-extraction is slightly better than object sub-extraction (see Table 5 for example materials). Then, the acceptability of subject and object sub-extraction is measured in comparison to various other constructions: an ungrammatical baseline, coordination, and full DP pied-piping (Abeillé et al., 2020).

| | |
|---|---|
| *Sub-Extraction* | |
| Object | The dealer sold a sportscar, of which the baseball player loved [the color __] because of its surprising luminance. |
| Subject | The dealer sold a sportscar, of which [the color __] delighted the baseball player because of its surprising luminance. |
| *Ungrammatical baseline (missing 'of')* | |
| Object | The dealer sold a sportscar, which the color __ the baseball player loved because of its surprising luminance. |
| Subject | The dealer sold a sportscar, which the color __ delighted the baseball player because of its surprising luminance. |
| *Coordination* | |
| Object | The dealer sold a sportscar, and the baseball player loved the color of the sportscar because of its surprising luminance. |
| Subject | The dealer sold a sportscar, and the color of the sportscar delighted the baseball player because of its surprising luminance. |
| *Full DP pied-piping* | |
| Object | The dealer sold a sportscar, [the color of which] the baseball player loved __ because of its surprising luminance. |
| Subject | The dealer sold a sportscar, [the color of which] __ delighted the baseball player because of its surprising luminance. |

**Table 5**. Materials from Abeillé et al. 2020, p. 10, examples 16a-h.

We argue that these comparisons do not properly account for differences between subject and object relativization if the goal is to isolate the cost of sub-extraction in each position. In particular, a direct comparison of subject and object sub-extraction in our Experiment 2 would not be suggestive of a "subject island effect" (like Abeillé et al., 2020), as both were rated as similarly degraded. However, as discussed in Section 3, we argue that measuring a subject island effect requires factoring out independent costs associated with the island violation, following Sprouse (2007) and Sprouse et al. (2012), and that these factors likely confound direct comparisons between subject and object sub-extraction. In particular, subject RCs are generally preferred over object RCs (for a review, see Lau & Tanaka, 2021), a finding which was replicated in our study. The contrast is assumed to reflect a processing difficulty that arises due to the longer dependency, and/or the intervening subject argument, in object RCs (King & Just, 1991; Gibson, 1998; Friedmann et al., 2009). Because these factors also apply to sub-extraction— as object sub-extraction results in a longer dependency which spans more intervening arguments than subject sub-extraction— they are expected to lower the acceptability rating of object sub-extraction, as well. Only when factoring out this baseline difference between subject and object relativization were we able to detect a subject island effect: The super-additive cost of subject sub-extraction was greater than that of object sub-extraction.

Abeillé et al. also compared sub-extraction to the three construction types in Table 5— an ungrammatical baseline (missing *of*), coordination and full DP pied-piping— and looked for interactions with position (subject/object). They report no significant interactions, i.e., no evidence for a greater difference between subjects and objects in the sub-extraction conditions compared to the other conditions. It is unclear, however, why these interactions should diagnose a subject island effect. First, the ungrammatical baseline ('missing *of*'), where a preposition is omitted, also involves sub-extraction from subject/object position, and thus a subject island effect should be reflected in its ratings as well. Second, the comparison between subjects and objects in the coordination baseline amounts to judging DP repetition in each position, which is expected to contrast based on discourse properties: the repeated antecedent is more accessible in the subject compared to the object position, potentially making its repetition less natural (Ariel, 1999). Crucially, in both cases, the conditions targeting subjects involve more predicted costs than the conditions targeting objects, and this could contribute to the lack of an interaction.

Lastly, the full DP pied-piping baseline introduces several challenges. This type of pied-piping, where an embedded DP pied-pipes its containing DP, has been argued to only be possible in non-restrictive RCs (22a), and unacceptable in other constructions involving movement, such as restrictive RCs (22b) or WHQs (22c).

(22) a. The dealer sold a sports car, [the color of which] the baseball player loved __.

b. *Out of all the sports cars, the dealer sold the sports car [the color of which] the baseball player loved __.

c. *I don't know [the color of which sports car] the baseball player loved __.

Pied-piping in non-restrictive RCs has been argued to be stylistically literary and marginal in acceptability, i.e., it is not a part of English speakers' colloquial speech (Emonds, 1976; Horvath, 2006; Webelhuth, 1992). Thus, it is unclear what participants' judgements for pied-piping in non-restrictive RCs reflect. In addition, research suggests that this type of pied-piping is further constrained, based on the category of the pied-piped phrase, its clausal position, and the location of the *wh*-phrase (Cable, 2010a, 2010b; Heck, 2008). These factors complicate the use of this construction as a baseline for subject versus object extraction.

We have shown that several independent and under-examined factors may influence the acceptability of the baseline constructions in Abeillé et al. (2020). In order to diagnose a subject island effect in RCs, it is necessary to factor in the fact that subject relativization is greatly preferred over object relativization, as this preference may obscure the cost of subject sub-extraction by independently lowering the ratings of object sub-extraction. This subject advantage was replicated in our experiments across the three constructions to varying degrees, highlighting its importance in cross-constructional comparisons. Moreover, the fact that the subject advantage varied across constructions— while the cost of sub-extraction from a subject was invariant (see Section 7)— provides evidence that these costs are independent, and motivates the use of designs that tease them apart.

### 8.2. Further predictions of *constructional IS profile* approaches to islands

In addition to the methodological considerations raised above, Abéille et al (2020)— and other accounts that attribute island effects to IS— make wrong predictions with regard to several findings widely known in the syntactic literature on islands. As Schütze et al. (2015) argue in response to Ambridge et al. (2014), proposals to reduce island constraints to independently motivated discourse factors must account for the broad range of findings on islands that have been documented in generative syntax. In particular, it is unclear how an account that attributes the subject island effect to IS factors could meaningfully distinguish those syntactic dependencies that are sensitive to subject islands from those which are not. Limiting the FBC to extraction raises a crucial question: if islands are solely due to IS clashes, why should only extraction dependencies, but not non-extraction dependencies with a similar use and meaning, be affected by IS clashes?

One such case is *wh* in-situ constructions. In some languages, WHQs do not involve overt extraction of the *wh*-phrase (e.g., Chinese, Korean). Instead, the *wh*-phrase must remain in its argument position (*wh* in-situ). Crucially, some *wh* in-situ phrases may appear within an island (Aoun & Li 1993; Huang, 1982). If these WHQs have the same interpretation as English ones, it remains unclear why the FBC should not cause the same degradation in acceptability (see Schütze et al., 2015 for relevant discussion concerning the argument/adjunct distinction in *wh* in-situ). A similar phenomenon is observed in languages like English, where the second *wh*-phrase in a multiple *wh*-question remains in-situ. Huang (1982) found that *wh* in-situ in multiple *wh*-questions is not sensitive to islands, including subject islands (23), corroborated by experimental work with a super-additive design (Sprouse et al., 2011).

(23) Who thinks the joke about what is funny?

The *wh* in-situ (*what*) is much like the moved *wh*-phrase (*who*), i.e., it requires an answer. Thus, it should introduce the same focus interpretation, and its presence within a subject should violate the FBC. Based on this finding, Winckel et al. (2025) propose a revised version of the FBC: "An extracted element should not be more focused than its (non-local) governor" (11) replacing the original: "A focused element should not be part of a backgrounded constituent." (10) (Abéille et al., 2020). While the original FBC was meant to apply broadly, the revised FBC in Winckel et al. (2025) is limited to extracted phrases.

The revision proposed in Winckel et. al. (2025) is necessary, but it is not trivial. The strength of the original FBC, and its initial appeal, was that it eliminated the need for a constraint on *movement*— an abstract mechanism that applies across discourse configurations— instead linking the subject island effect to a discourse-based notion: *focusing*. Unless there is a meaningful discourse-based distinction between focusing in general, and focusing when it applies to an extracted element, it is unclear why only the latter should be sensitive to the FBC.

In addition to *wh* in-situ, there is further support that island violations are defined by the presence of a movement dependency across a particular syntactic domain, rather than focusing a backgrounded constituent. As we have mentioned above, when the subject internal gap is parasitic to a gap in the matrix clause, its acceptability greatly improves (as first identified by Ross, 1967, and experimentally corroborated by Phillips, 2006).

Similarly, there are languages in which island violations are improved when the dependency is resolved by a pronoun instead of a gap, called a resumptive pronoun (RP). In languages that have grammatical RPs (e.g., Irish, varieties of Arabic, Hebrew), their presence has been reported to ameliorate the acceptability of island violations, both observationally and experimentally (Sells, 1987; McCloskey, 2006; Keshev & Meltzer Asscher, 2017; Tucker et al. 2019). Thus, for example, while a gap within a subject is ungrammatical in Hebrew, the same dependency is grammatical when resolved by a RP (24) (Sichel, 2022).

(24) ze ha-iš še-[laxšov še-tifgeši *__ / oto] yihiye tipši

this the-man that to.think that-you.FUT.meet him would.be silly

'This is the man that to think that you'd meet him would be silly.' (Sichel, 2022)

Like PGs, it is unclear how a RP could overcome the FBC violation in subject sub-extraction. On the other hand, RPs' insensitivity to islands, along with other properties, follows from a syntactic analysis of resumed dependencies as not involving extraction (Chomsky, 1977; Borer, 1984; McCloskey, 1990; Sichel, 2022). This, again, suggests that what is sensitive to island locality is a movement dependency, not an IS profile.

In addition, Abéille et al. (2020), based on the original FBC, suggest that alongside a restriction on WHQs, subjects should also resist *prosodic* focusing of a sub-constituent, as this would similarly introduce an IS clash. As support for this intuition, they provide a marginal ("?") judgement for (25).

(25) ? the color *of the blue car* delighted the football player

However, the judgment provided for example (25) does not seem to be shared across a large sample of English speakers; the speakers which we informally surveyed all accept it. An account like the original FBC predicts (25) to be as unacceptable as bona fide subject island violations, as they both introduce the same IS clash, but this does not seem well-supported.

Given the revised FBC proposed in Winckel et al. (2025), which is rearticulated as a ban on IS clashes in an extraction dependency, as well as the phenomena discussed above, it is clear that the constraint responsible for subject island effects must include reference to extraction— a syntactic notion— and that subject island effects cannot be brushed off as a general tension between conflicting IS demands.

### 8.3. IS may still play a role in island effects

Abeillé et. al. (2020)'s proposal relies on the assumption that the categories 'subject' and 'object', and how they interact with the possibility of sub-extraction, directly map onto a contrast in discourse-backgroundedness. However, 'subject' and 'object' are not monolithic categories. Both subjects and objects may vary in thematic role, topicality, definiteness, animacy, and syntactic position, and these properties can meaningfully affect their discourse status as well as the possibility for sub-extraction. Though we argue against the *constructional IS profile* approach to islandhood, we do not rule out the possibility that notions such as backgroundedness may play a role in the characterization of islands (cf. Momma & Dillon, 2023), or that distinct dependencies may be differentially impacted by constraints on locality, understood syntactically (Kush et. al., 2018, 2019; Kobzeva et al., 2022).

Although an IS clash between the filler and extraction domain does not seem to have a big impact on subject island effects, other IS properties could, such as the presuppositional nature of the subject DP. Presuppositionality of a DP has been identified as a relevant factor for the relative acceptability of sub-extraction from object DPs in German (Diesing, 1992), subject DPs in English (Chung & McCloskey, 1983; McCawley, 1981), object DPs in English (Fiengo & Higginbotham, 1981), subject DPs in Italian (Bianchi & Chesi, 2014), and in the investigation of a different purported syntactic island: RCs[4]. Researchers have found that the well-formedness of extraction out of RCs is mediated by the IS of the DP that hosts the RC: island effects only arise when sub-extraction occurs out of a presuppositional DP— a 'given' DP, in Abeillé et al.'s terminology (Lindahl, 2014, 2017; Nyvad et al., 2017; Sichel, 2018; Vincent et al., 2022).

A number of theories have linked presuppositionality to the position of a subject (Diesing, 1992; Sichel, 2018; Bianchi & Chesi, 2014), suggesting that presupposed DPs have undergone movement to a high syntactic position in contrast to non-presupposed DPs, which have not moved from their base position. Under this perspective, subjects are anticipated to disallow sub-extraction when the subject maps to a high syntactic position (which is true for all of the subjects in both ours and Abeillé et. al. 2020's stimuli), but permeable to sub-extraction when it maps to a low syntactic position. The IS property of presuppositionality correlates with these different positions.

[4] Distinct from *relativization*, the filler-gap dependency investigated in Experiment 2 of this paper, this island indicates that relativized constructions are themselves opaque to any further instances of sub-extraction. This phenomenon falls under the umbrella term of "Complex NP Islands" in syntactic literature, which refers to the family of islands that all involve a domain of extraction that is a noun phrase which contains an embedded clause (Ross, 1967; Adger, 2003). Though we focus the discussion in 8.3 on DP islands and Complex NP Island phenomena, we note that other researchers have appealed to concepts similar to presuppositionality for other kinds of islands as well (e.g., C.-T. J. Huang, 1982; Kush et al., 2018; N. Huang et al., 2025).

The empirical observations coming out of all of these works highlight that the definition of islandhood, when it comes to DPs, depends on finer-grained syntactic properties than the grammatical functions 'subject' and 'object'. These accounts, just as *constructional IS profile* and *direct backgroundedness* approaches, agree that information structural factors are important, but differ in how information structure is mapped to syntax, as well as the significance of syntactic position vs. grammatical function in the mapping between syntax and information structure: within the class of subjects and within the class of objects, 'given' DPs are mapped to higher positions. In this sense, our findings are compatible with the possibility that IS may play some role in the acceptability of sub-extraction, but we suspect that role is likely indirect, mediated by syntactic position.

## 9. Conclusion

This study investigated the degree of abstractness of syntactic constraints: Can all restrictions on the form of linguistic expressions be attributed to aspects of language use and meaning, or is there reason to believe that language users possess a dedicated system of syntactic rules, autonomous of language's communicative function? Specifically, we focused on the highly debated subject island constraint. We compared the hypothesis that the subject island constraint is the result of a pressure to avoid information structure (IS) "clashes" (Abeillé et al., 2020; Winckel et al., 2025) with syntactic approaches that assume this constraint applies to an abstract movement dependency, independent of construction-specific discourse function. Three large scale acceptability studies probed for a subject island effect in three constructions which differ in their IS profile: topicalization (TOP), wh-questions (WHQ), and relative clauses (RC). We observed degraded acceptability of sub-extraction from subjects vs. objects across the construction types that were predicted to modulate the presence of a subject island violation under discourse-based accounts of islandhood. Our results indicate the ban on subject sub-extraction cannot be reduced to discourse-based notions, and is best attributed to the dependency uniting those constructions: movement.

**Data availability**

For access to all experiment materials, including itemsets, fillers, experiment code, results, and analysis files, please visit our OSF data repository: osf.io/pmbd9/overview.

**Acknowledgements**

We thank Lisa Pham, Alison Sun, and Matthew Vasser for their assistance with material creation and data collection. We are also grateful to Jake Vincent for helpful discussion and methodological resources. Lastly, we thank the attendees of the 30th Architectures and Mechanisms for Language Processing (AMLaP), the 38th Annual Conference on Human Sentence Processing (HSP), and the 47th Annual Meeting of Generative Linguistics in the Old World (GLOW) for feedback on earlier versions of this work.

**Funding**

This research was supported by NSF BCS Award #2019804 to UC Santa Cruz (M. Toosarvandani, PI; I. Sichel, M. Wagers, Co-PIs).